\documentclass[letterpaper]{article} % DO NOT CHANGE THIS
\usepackage{aaai2027}  % DO NOT CHANGE THIS
\nocopyright
\usepackage[hyphens]{url}  % DO NOT CHANGE THIS
\usepackage{graphicx} % DO NOT CHANGE THIS
\usepackage{natbib}  % DO NOT CHANGE THIS AND DO NOT ADD ANY OPTIONS TO IT
\usepackage{caption} % DO NOT CHANGE THIS AND DO NOT ADD ANY OPTIONS TO IT
\usepackage{algorithm}
\usepackage{algorithmic}
\usepackage[table]{xcolor}
\usepackage{colortbl}
\usepackage{nicematrix}
\usepackage{booktabs}
\usepackage{multirow}
\usepackage{amsmath}
\usepackage{amssymb}
\usepackage{tabularx}

\usepackage[hyphens]{url}  % DO NOT CHANGE THIS
\usepackage{graphicx} % DO NOT CHANGE THIS
\usepackage{natbib}  % DO NOT CHANGE THIS AND DO NOT ADD ANY OPTIONS TO IT
\usepackage{caption} % DO NOT CHANGE THIS AND DO NOT ADD ANY OPTIONS TO IT
\usepackage{algorithm}
\usepackage{algorithmic}
\usepackage[table]{xcolor}
\usepackage{colortbl}
\usepackage{nicematrix}
\usepackage{booktabs}
\usepackage{multirow}
\usepackage[most]{tcolorbox}
\usepackage{amsmath}
\usepackage{amssymb}
\usepackage{longtable}
\usepackage{newfloat}
\usepackage{listings}

\usepackage{newfloat}
\usepackage{listings}
\DeclareCaptionStyle{ruled}{labelfont=normalfont,labelsep=colon,strut=off} % DO NOT CHANGE THIS
\floatstyle{ruled}
\newfloat{listing}{tb}{lst}{}
\floatname{listing}{Listing}

\usepackage{booktabs}

\title{Beyond Confidence: Stability-Aware Test-Time Adaptation for LLM Reasoning}
\author{
    Written by AAAI Press Staff\textsuperscript{\rm 1}\thanks{With help from the AAAI Publications Committee.}\\
    AAAI Style Contributions by Peter Patel Schneider,
    Sunil Issar,\\
    J. Scott Penberthy,
    George Ferguson,
    Hans Guesgen,
    Francisco Cruz\equalcontrib\corresponding,
    Marc Pujol-Gonzalez\equalcontrib\corresponding
}
\affiliations{
    \textsuperscript{\rm 1}Association for the Advancement of Artificial Intelligence\\
    1101 Pennsylvania Ave, NW Suite 300\\
    Washington, DC 20004 USA\\
    proceedings-questions@aaai.org
}

\title{Beyond Confidence: Stability-Aware Test-Time Adaptation for LLM Reasoning}
\author {
    Bincheng Gu\textsuperscript{\rm 1},
    Min Gao\textsuperscript{\rm 1}\corresponding,
    Zongwei Wang\textsuperscript{\rm 1},
    Yibing Bai\textsuperscript{\rm 1},
    Yulan He\textsuperscript{\rm 2},
    Junliang Yu\textsuperscript{\rm 3}
}
\affiliations {
    \textsuperscript{\rm 1}Key Laboratory of Dependable Service Computing in Cyber Physical Society, Chongqing
University, China\\
    \textsuperscript{\rm 2}King’s College London, London, UK \\
    \textsuperscript{\rm 3}Griffith University, Brisbane, Australia\\
    gaomin@cqu.edu.cn, 
}

\begin{document}

\maketitle

\begin{abstract}
Test-time adaptation has emerged as a lightweight alternative to costly post-training for improving the reasoning capabilities of Large Language Models (LLMs) on downstream tasks. Predictive entropy provides a model-derived signal for such adaptation, guiding models toward higher-confidence reasoning states without external verifiers or reward models. However, higher confidence does not necessarily imply correctness, as LLMs may remain highly confident along incorrect reasoning trajectories. We observe that high-confidence reasoning is more likely to be correct when confidence remains stable under local perturbations. Based on this observation, we propose \underline{T}est-Time \underline{A}daptation via \underline{S}tability-Aware \underline{C}onfidence \underline{O}ptimization (TASCO), a framework that incorporates local stability into confidence-based test-time adaptation while keeping the LLM frozen. TASCO operationalizes local stability by optimizing a lightweight task-level prefix under two alternative perturbation strategies: Random Perturbation promotes distributional stability across trajectories induced by nearby perturbed prefixes, whereas Sharpness-Aware Perturbation targets worst-case local sensitivity. Experiments demonstrate that TASCO improves reasoning accuracy and token efficiency across diverse LLMs and reasoning benchmarks, while behavioral analyses show that it maintains stable confidence under local perturbations without prematurely concentrating the model’s predictive distribution.
\end{abstract}

\section{Introduction}

% Large Language Models (LLMs) have demonstrated strong reasoning capabilities on complex tasks such as mathematical reasoning and code generation~\cite{guo2025deepseek,jaech2024openai}. However, adapting these pretrained models to specific downstream tasks has traditionally relied on post-training methods that update model parameters, such as supervised fine-tuning or reinforcement learning~\cite{xu2025toward,lambert2024tulu,muennighoff2025s1}. Although effective, these methods often require substantial computational resources. This cost has motivated recent work to keep model parameters frozen and instead optimize continuous inputs or internal representations at inference time~\cite{li2023inference}.

Large Language Models (LLMs) have demonstrated strong reasoning capabilities on complex tasks such as mathematical reasoning and code generation~\cite{guo2025deepseek,jaech2024openai}. Traditionally, adapting these models to downstream tasks relies on post-training methods that update model parameters, such as supervised fine-tuning or reinforcement learning~\cite{shao2024deepseekmath,lambert2024tulu}. Although effective, these methods require substantial computational resources, motivating recent work to keep model parameters frozen and instead optimize continuous inputs or internal representations at inference time~\cite{hu2025slot}.

One effective paradigm is to optimize a small set of continuous variables, including soft prompts, steering vectors, and latent representations, to guide a frozen LLM toward task-specific reasoning states~\cite{softcot,latentseek,sinii2025steering}. These approaches differ mainly in the source of their optimization signals. Some rely on external verifiers or reward models to score answers or reasoning trajectories~\cite{khalifa2025process,nguyen2026atlas}, whereas verifier-free methods use the model's output entropy as an intrinsic confidence signal~\cite{ye2025ltpo}.

% However, entropy-based test-time optimization assumes that an LLM is more likely to answer correctly when its predictive entropy is lower during reasoning~\cite{agarwal2026unreasonable}. This assumption can fail on complex reasoning tasks, where LLMs can be poorly calibrated and remain highly confident along incorrect reasoning trajectories~\cite{xiong2024can,wang2024self}. When confidence is the only test-time objective, entropy minimization cannot distinguish high confidence associated with reliable reasoning from spurious confidence along an incorrect trajectory, and may reinforce the latter \cite{chen2026zerosiam,zhao2026echo}.

However, entropy-based test-time optimization assumes that an LLM is more likely to answer correctly when its predictive entropy is lower during reasoning~\cite{agarwal2026unreasonable}. This assumption can fail on complex reasoning tasks, where LLMs can be poorly calibrated and remain highly confident along incorrect reasoning trajectories~\cite{jurayj2025your, huang2025efficient}. When confidence is the only test-time objective, entropy minimization cannot distinguish high confidence associated with reliable reasoning from spurious confidence along an incorrect trajectory, potentially steering the model toward a locally fragile high-confidence state~\cite{chen2026zerosiam,zhao2026echo}.

\begin{figure}[t]
    \centering
    \includegraphics[width=0.48\textwidth]{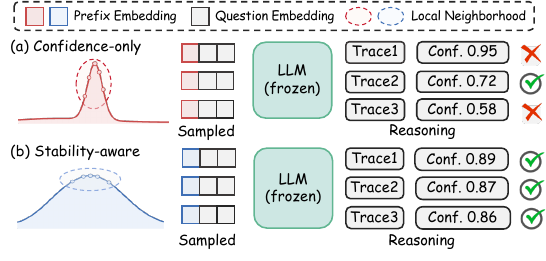}
    \caption{Comparison of confidence-only and stability-aware optimization under prefix perturbations across trajectories yielding correct and incorrect answers.}
    \label{fig:intro1}
\end{figure}

% How can we guide an LLM toward a high-confidence state that supports reliable reasoning? Inspired by the connection between flat minima and generalization~\cite{keskar2016large}, we examine local stability as a potential indicator of reasoning reliability. Specifically, high confidence may better reflect reliability when maintained across a local neighborhood of the steering variable rather than concentrated in a narrow peak (Fig.~\ref{fig:intro1}). To examine this idea, we perturb the steering variable and quantify local stability using the variance of trajectory-level confidence. Empirically, lower variance is associated with higher answer accuracy, and this association remains significant after accounting for mean confidence and persists among high-confidence queries (see Appendix~\ref{app:perturbation_analysis}). Together, these findings suggest that confidence magnitude alone does not fully characterize reliability and that local stability provides a complementary label-free signal.

\begin{figure}[t]
    \centering
    \includegraphics[width=0.47\textwidth]{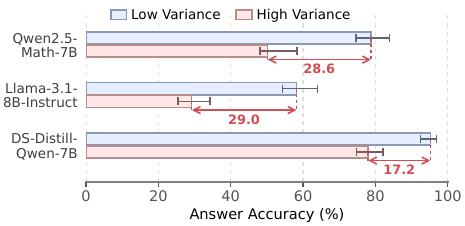}
    \caption{Answer accuracy for low- and high-variance query groups defined by the median trajectory-confidence variance.}
    \label{fig:intro2}
\end{figure}

How can we guide an LLM toward a high-confidence state that supports reliable reasoning? Inspired by the connection between flat minima and generalization~\cite{keskar2016large}, we examine the local stability of high-confidence states. Specifically, high confidence that remains stable under local perturbations to the steering variable is more likely to stem from reliable reasoning than from a fragile overconfident state (Fig.~\ref{fig:intro1}). We measure this stability using trajectory-level confidence variance. As shown in Fig.~\ref{fig:intro2}, the low-variance group consistently achieves higher answer accuracy than the high-variance group across representative models from three families. Further analysis shows that the low-variance group remains more accurate among queries with comparable mean confidence (see Appendix B for details). Together, these results suggest that local stability complements confidence magnitude in characterizing reasoning reliability.

% How can we guide an LLM toward a high-confidence state that supports reliable reasoning? Inspired by the connection between flat minima and generalization~\cite{keskar2016large}, we examine the local stability of high-confidence reasoning states. Specifically, confidence may be more informative when maintained across a local neighborhood of the steering variable rather than concentrated in a narrow peak (Fig.~\ref{fig:intro1}). We quantify local stability by perturbing the steering variable and measuring the variance of trajectory-level confidence. As shown in Fig.~\ref{fig:intro2}, lower variance is consistently associated with higher unperturbed answer accuracy across three representative model families. The same pattern holds among queries with comparable confidence and remains evident on difficult queries (see Appendix~\ref{app:perturbation_analysis} for details). Together, these findings motivate local stability as a complementary signal for assessing high-confidence reasoning.

% \underline{T}est-Time \underline{A}daptation via \underline{S}tability-Aware \underline{C}onfidence \underline{O}ptimization (TASCO)

% gu
Guided by this observation, we propose Test-Time Adaptation via Stability-Aware Confidence Optimization (TASCO), a label-free framework for guiding frozen LLMs toward reliable high-confidence reasoning. Concretely, TASCO optimizes a task-level prefix shared across unlabeled test queries to stabilize confidence during reasoning. Once optimized, this prefix is prepended to each query embedding, thereby steering the frozen model throughout trace generation. Since local stability of reasoning confidence can manifest as either variation across nearby perturbed prefixes or sensitivity to the most adversely perturbed prefix, TASCO operationalizes it using two complementary strategies: Random Perturbation measures distributional stability through confidence variation across trajectories independently decoded under Gaussian-perturbed prefixes, while Sharpness-Aware Perturbation targets worst-case local sensitivity by applying a single gradient-guided perturbation along the most sensitive direction in the prefix space for the current trajectory. By optimizing the prefix with these two schemes, TASCO improves accuracy by up to 17.2\% while generating up to 28.1\% fewer tokens across diverse LLMs and reasoning benchmarks. Behavioral analyses show TASCO promotes stable confidence under local perturbations without prematurely narrowing reasoning directions. The learned prefix improves reasoning on unseen datasets without further optimization.

%gao
% Guided by this observation, we propose Test-Time Adaptation via Stability-Aware Confidence Optimization (TASCO), a label-free framework for guiding frozen LLMs toward reliable high-confidence reasoning. TASCO optimizes a task-level prefix shared across unlabeled test queries so that model confidence remains stable under nearby prefix perturbations. However, translating this observation into an optimization objective requires determining how local stability should be characterized. A reliable high-confidence state should exhibit consistent confidence across typical nearby conditions and avoid sharp changes along the most sensitive local direction. This motivates TASCO to characterize local stability from distributional and worst-case perspectives through two alternative strategies. Random Perturbation measures distributional stability through confidence variation across trajectories independently decoded under Gaussian-perturbed prefixes, whereas Sharpness-Aware Perturbation probes worst-case stability along the most sensitive prefix direction for the current trajectory using a single gradient-guided perturbation. Across diverse LLMs and reasoning benchmarks, TASCO improves accuracy by up to 18.8\% while generating up to 28.1\% fewer tokens. Behavioral analyses show TASCO promotes stable confidence under local perturbations without prematurely narrowing reasoning directions. The learned prefix improves reasoning on unseen datasets without further optimization.

Our contributions are summarized as follows:
\begin{itemize}
    \item We introduce a stability-aware perspective on confidence-driven test-time optimization, identifying local stability as a complementary signal for assessing the reliability of high-confidence reasoning.

    \item We propose TASCO, a label-free framework that uses two alternative perturbation strategies to guide frozen LLMs toward high-confidence states that remain stable under local perturbations. We further characterize how both strategies control local prefix sensitivity.
    
    \item Extensive experiments across diverse LLMs and reasoning benchmarks demonstrate that TASCO improves reasoning accuracy and efficiency while promoting stable confidence under local perturbations.
\end{itemize}

\section{Related Work}

\subsection{Reasoning in Large Language Models}

Chain-of-thought (CoT) prompting~\cite{wei2022chain} improves LLM reasoning by eliciting explicit intermediate steps. Building on this paradigm, subsequent methods enhance reasoning through trajectory aggregation~\cite{wang2022self}, structured search~\cite{yao2023tree}, and step-level verification~\cite{zhang2025enhancing}. Reasoning models further extend explicit CoT by performing longer inference-time deliberation~\cite{chen2026towards}. In parallel, continuous reasoning methods move part of the intermediate computation from token space into latent space. COCONUT~\cite{coconut} and CODI~\cite{shen2025codi}, for example, replace explicit reasoning tokens with latent states, while SoftCoT~\cite{softcot} and SemCoT~\cite{he2026semcot} use auxiliary modules to construct continuous rationales that are subsequently aligned with and injected into the target model~\cite{wei2025sim}. Together, these methods establish latent states as an interface for steering reasoning without backbone updates.

\subsection{Test-Time Adaptation for Reasoning}

Test-time adaptation improves LLM reasoning using inference-time signals. TTRL~\cite{zuo2026ttrl} derives majority-vote pseudo-rewards from rollouts. Entropy-based approaches either use negative token entropy for policy optimization~\cite{prabhudesai2025maximizing} or minimize output entropy to adapt model parameters on unlabeled samples~\cite{gao2025one}. Such updates add computation and risk catastrophic forgetting~\cite{hu2025test}. Recent methods instead optimize continuous variables while freezing the model. Some rely on external verifiers or reward models to guide intermediate reasoning states~\cite{han2026steer2adapt,chen2025seal}, whereas others use model-derived entropy to optimize continuous vectors~\cite{kang2026model,ye2025ltpo}. MTI selectively guides high-entropy tokens during decoding~\cite{yang2026less}. However, confidence-driven methods rarely assess whether the resulting states are reliable, despite evidence that LLM confidence can be misaligned with correctness~\cite{agarwal2026unreasonable}. Perturbation-based studies instead use sensitivity diagnostically: CCPS predicts answer correctness from changes in hidden representations~\cite{khanmohammadi2025calibrating}, while Wen et al. identify uncertain reasoning steps through embedding perturbations~\cite{wen2026embedding}. Unlike these approaches, TASCO optimizes local stability to promote reliable reasoning.

\begin{figure*}[t]
    \centering
    \includegraphics[width=\textwidth]{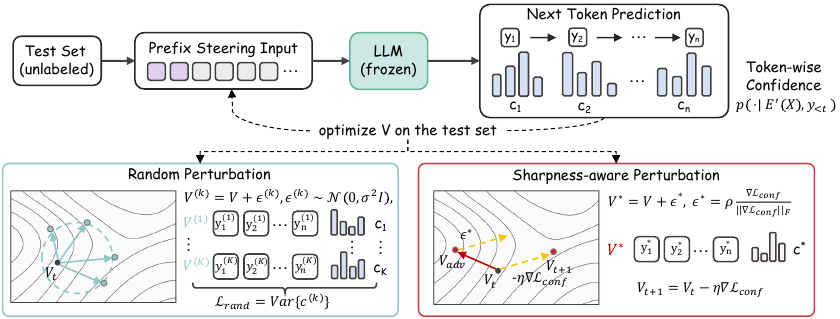}
    \caption{Overview of TASCO. TASCO optimizes a shared prefix for confidence and stability. Random Perturbation penalizes local confidence variance, while Sharpness-Aware Perturbation targets worst-case sensitivity.}
    \label{fig:framework}
\end{figure*}

\section{Method}

% In this section, we present TASCO for optimizing a continuous prefix at test time while keeping the LLM frozen. We define its confidence objective, introduce two perturbation-based strategies, and theoretically analyze their effects on local prefix sensitivity. Fig.~\ref{fig:framework} provides an overview of TASCO. 

In this section, we present TASCO, a stability-aware framework that optimizes a continuous prefix at test time while keeping the LLM frozen. We first formulate the optimization problem and define the confidence objective. We then introduce two alternative strategies for characterizing local stability from distributional and worst-case perspectives. We further analyze how these strategies guide confidence optimization. Figure~\ref{fig:framework} provides an overview of TASCO.

\subsection{Preliminaries}
\paragraph{Problem Formulation.}
Let $M_\theta$ denote a pretrained LLM with frozen parameters $\theta$, and let $\mathcal{D}_{\mathrm{test}}=\{X_1,\dots,X_N\}$ be an unlabeled test set. Our goal is to improve the reasoning performance of $M_\theta$ on $\mathcal{D}_{\mathrm{test}}$ without updating its parameters. To this end, we optimize a shared continuous prefix $V\in\mathbb{R}^{L\times d}$. Here, $L$ and $d$ denote the prefix length and embedding dimension. Given a query $X$ with token embeddings $E(X)$, we prepend $V$ to form the prefix-conditioned input $E'(X)=[V;E(X)]$, using $[\cdot;\cdot]$ to denote concatenation along the sequence dimension. The frozen model then autoregressively generates a reasoning trajectory $Y=(y_1,\dots,y_T)$. At decoding step $t$, its next-token distribution is $p_\theta(\cdot\mid E'(X),y_{<t})$. TASCO optimizes $V$ using only label-free signals derived from $M_\theta$, without external supervision or auxiliary models.

\paragraph{Confidence Objective.}
We use predictive entropy as a label-free confidence signal. For a generated trajectory $Y=(y_1,\dots,y_T)$, the confidence loss is defined as the average token-level entropy:
\begin{equation}
\mathcal{L}_{\mathrm{conf}}(V;X,Y)=\frac{1}{T}\sum_{t=1}^{T}
H\left(p_\theta(\cdot\mid E'(X),y_{<t})\right),
\label{eq:confidence_loss}
\end{equation}
where $H(p)=-\sum_{w\in\mathcal{V}}p(w)\log p(w)$ denotes the entropy over the vocabulary $\mathcal{V}$. In practice, we compute this loss using one generated trajectory per input and treat the generated tokens as fixed.

% \subsection{Perturbation-Based Optimization}

% While entropy minimization encourages confident reasoning, confidence alone does not guarantee correctness. Appendix~\ref{app:perturbation_analysis} shows that reasoning accuracy is associated with confidence stability under small prefix perturbations. We therefore extend confidence optimization with two perturbation-based strategies: (1) Random Perturbation and (2) Sharpness-Aware Perturbation.

\subsection{Confidence Optimization with Local Stability}

Although entropy minimization offers a label-free objective for prefix optimization, it may steer the prefix toward a locally fragile high-confidence state. TASCO therefore introduces two local-stability strategies: (1) Random Perturbation for distributional stability and (2) Sharpness-Aware Perturbation for worst-case directional stability.

\subsubsection{Random Perturbation.}

We promote distributional stability by examining reasoning trajectories decoded under nearby prefixes. At each optimization step, we draw $K$ Gaussian perturbations around the current prefix:
\begin{equation}
V^{(k)} = V + \epsilon^{(k)}, \quad
\epsilon^{(k)} \sim \mathcal{N}(0,\sigma^2 I),
\label{eq:random_prefix}
\end{equation}
where $\sigma$ is the standard deviation of the Gaussian noise applied to each prefix element. For each input $X$, the model greedily decodes a trajectory $Y^{(k)}$ under every perturbed prefix $V^{(k)}$. We summarize each rollout by the average log-likelihood of its generated tokens:
\begin{equation}
c^{(k)}=\frac{1}{T_k}\sum_{t=1}^{T_k}\log p_\theta
\left(y_t^{(k)}\mid E'^{(k)}(X), y_{<t}^{(k)}\right).
\label{eq:trajectory_confidence}
\end{equation}

Variation in $\{c^{(k)}\}_{k=1}^{K}$ reflects local behavioral instability and may capture trajectory switching because each perturbed prefix is re-decoded. We quantify the behavioral sharpness of $V$ on $X$ by the sample variance:
\begin{equation}
\mathcal{S}(V;X)=\frac{1}{K-1}\sum_{k=1}^{K}\left(c^{(k)}-\bar{c}\right)^2,\quad
\bar{c}=\frac{1}{K}\sum_{k=1}^{K}c^{(k)}.
\label{eq:random_sharpness}
\end{equation}
For a mini-batch $\mathcal{B}$, the random-perturbation regularizer is obtained by averaging this quantity across inputs,
$\mathcal{L}_{\mathrm{rand}}(V)=
|\mathcal{B}|^{-1}\sum_{X\in\mathcal{B}}\mathcal{S}(V;X)$.
Combining it with the batch-averaged confidence loss gives:
\begin{equation}
\mathcal{L}_{\mathrm{RP}}=\mathcal{L}_{\mathrm{conf}}(V)+ \lambda_{\mathrm{rand}}\mathcal{L}_{\mathrm{rand}}(V),
\label{eq:random_objective}
\end{equation}
where $\lambda_{\mathrm{rand}}$ balances predictive confidence and local stability. Thus, $\mathcal{L}_{\mathrm{conf}}$ encourages low predictive entropy, whereas $\mathcal{L}_{\mathrm{rand}}$ reduces behavioral variation across nearby prefixes.

\subsubsection{Sharpness-Aware Perturbation.}

Unlike Random Perturbation, Sharpness-Aware Perturbation avoids generating multiple trajectories by constructing a single gradient-guided perturbation. At each update, we fix the trajectory $Y$ decoded under $V$ while minimizing the worst-case confidence loss:
\begin{equation}
\min_V \max_{\|\epsilon\|_F \leq \rho}
\mathcal{L}_{\mathrm{conf}}(V+\epsilon;X,Y),
\label{eq:sap_minmax}
\end{equation}
where $\rho$ defines the perturbation radius under the Frobenius norm. Since the inner maximization in Eq.~\ref{eq:sap_minmax} is generally intractable, we use a first-order approximation of the confidence loss around $V$, based on
\begin{equation}
g=\nabla_V\mathcal{L}_{\mathrm{conf}}(V;X,Y).
\label{eq:sap_gradient}
\end{equation}

The linearized inner problem yields the following worst-case perturbation and perturbed prefix:
\begin{equation}
\widehat{V}=V+\epsilon^\star,\quad
\epsilon^\star=\rho\frac{g}{\|g\|_F},
\label{eq:sap_perturbation}
\end{equation}
where $\epsilon^\star$ is the perturbation along the steepest ascent direction under $\|\epsilon\|_F\leq\rho$. We then evaluate the confidence loss at $\widehat{V}$ using the same trajectory $Y$:
\begin{equation}
\mathcal{L}_{\mathrm{SAP}}=\mathcal{L}_{\mathrm{conf}}(\widehat{V};X,Y).
\label{eq:sap_loss}
\end{equation}
When updating $V$ with $\mathcal{L}_{\mathrm{SAP}}$, we treat $\epsilon^\star$ as constant to avoid second-order derivatives through $g$.

The variants trade off neighborhood coverage and computation: Random Perturbation captures variation across sampled perturbations, whereas SAP targets worst-case sensitivity with one gradient-guided perturbation. Complete algorithms appear in Appendix D.

\subsection{Theoretical Analysis}

We analyze how the two objectives control local confidence sensitivity to prefix perturbations. Random Perturbation captures confidence variation across perturbed rollouts, whereas SAP targets worst-case sensitivity along a fixed trajectory.

\subsubsection{Random Perturbation}
For $\epsilon\sim\mathcal{N}(0,\sigma^2I)$, let $Z$ denote the greedy trajectory decoded under $V+\epsilon$, and let $C=c(V+\epsilon)$ denote its trajectory-level confidence. The sample variance in Eq.~\ref{eq:random_sharpness} is an unbiased estimator of the following population behavioral sharpness:
\begin{equation}
\mathcal{S}_\sigma(V;X)
=\operatorname{Var}_{\epsilon\sim\mathcal{N}(0,\sigma^2I)}
\left[c(V+\epsilon)\right].
\label{eq:population_sharpness}
\end{equation}
Because each perturbed prefix is independently re-decoded, the law of total variance gives:
\begin{equation}
\begin{aligned}
\mathcal{S}_\sigma(V;X)
={}&
\mathbb{E}_{Z}\!\left[\operatorname{Var}(C\mid Z)\right]
+
\operatorname{Var}_{Z}\!\left(\mathbb{E}[C\mid Z]\right).
\end{aligned}
\label{eq:variance_decomposition}
\end{equation}
The first term captures variation among perturbations yielding the same trajectory, whereas the second captures variation in mean confidence across decoded trajectories, without differentiating through decoding.

For the within-trajectory term, consider a local neighborhood in which greedy decoding remains $Y$, and let $f_Y(V)$ denote the teacher-forced confidence of $Y$ in Eq.~\ref{eq:trajectory_confidence}. If $f_Y$ is sufficiently smooth, then, as $\sigma\to0$:
\begin{equation}
\operatorname{Var}_{\epsilon}[f_Y(V+\epsilon)]=\sigma^2\|\nabla_V f_Y(V)\|_F^2+
\mathcal{O}(\sigma^4).
\label{eq:random_first_order}
\end{equation}
Thus, away from stationary points, Random Perturbation penalizes first-order confidence sensitivity to isotropic perturbations. At a stationary point of $f_Y$, the corresponding expansion is:
\begin{equation}
\operatorname{Var}_{\epsilon}[f_Y(V+\epsilon)]=\frac{\sigma^4}{2}\|\nabla_V^2 f_Y(V)\|_F^2+o(\sigma^4).
\label{eq:random_second_order}
\end{equation}
The latter reflects aggregate local curvature, while Eq.~\ref{eq:variance_decomposition} additionally captures between-trajectory variation.

\subsubsection{Sharpness-Aware Perturbation}
For a trajectory $Y$ held fixed within an update, define
$\ell(V)=\mathcal{L}_{\mathrm{conf}}(V;X,Y)$. Its worst-case local increase over a Frobenius $\rho$-ball is:
\begin{equation}
\Delta_\rho(V)
=\max_{\|\epsilon\|_F\leq\rho}
\left[\ell(V+\epsilon)-\ell(V)\right].
\label{eq:sap_local_increase}
\end{equation}
Assume that $\ell$ has a $\beta$-Lipschitz-continuous gradient on this ball, and let $g=\nabla_V\ell(V)$. Then:
\begin{equation}
\left|\Delta_\rho(V)-\rho\|g\|_F\right|\leq\frac{\beta\rho^2}{2}.
\label{eq:sap_sensitivity_bound}
\end{equation}
Thus, the gradient norm provides a first-order approximation to the worst-case local increase. For $g\neq0$, the perturbation $\epsilon^\star$ in Eq.~\ref{eq:sap_perturbation} satisfies:
\begin{equation}
\left|\ell(V+\epsilon^\star)-\ell(V)-\rho\|g\|_F
\right|\leq\frac{\beta\rho^2}{2}.
\label{eq:sap_gradient_bound}
\end{equation}
Together, Eqs.~\ref{eq:sap_sensitivity_bound} and~\ref{eq:sap_gradient_bound} show that the induced increase $\ell(V+\epsilon^\star)-\ell(V)$ approximates $\Delta_\rho(V)$ with $\mathcal{O}(\rho^2)$ error.

Separately, if $V$ is a stationary point and $\ell$ is twice continuously differentiable in a neighborhood of $V$, the ideal inner problem admits the following expansion as $\rho\to0$:
\begin{equation}
\Delta_\rho(V)=\frac{\rho^2}{2}
\left[\lambda_{\max}\!\left(\nabla_V^2\ell(V)\right)\right]_++o(\rho^2),
\label{eq:sap_second_order}
\end{equation}
where $[a]_+=\max(a,0)$. Thus, the leading term of the ideal worst-case increase is governed by the largest positive local curvature.

Together, the two objectives control local confidence sensitivity under sampled and worst-case perturbations. These results characterize robustness rather than correctness; the empirical relationship between stability and answer accuracy is examined in Appendix B.

\section{Experiments}

\begin{table*}[t]
\centering
\small
\renewcommand{\arraystretch}{0.95}
\begin{NiceTabular}{llccccc|cc}
\toprule
\textbf{Model}
& \textbf{Method}
& \textbf{MATH500}
& \textbf{AMC23}
& \textbf{Minerva}
& \textbf{AIME24}
& \textbf{GPQA}
& \textbf{Avg.}
& \textbf{$\Delta$} \\
\midrule

\multirow{7}{*}{Qwen2.5-Math-1.5B}
& Zero-Shot CoT
& 39.4 & 34.7 & 8.8 & 6.7 & 20.2 & 22.0 & -- \\
& SLOT
& 43.4 & 47.2 & 16.9 & 10.0 & \underline{26.8} & 28.9 & +6.9 \\
& LatentSeek
& 45.6 & 43.4 & 14.3 & 7.5 & 22.2 & 26.6 & +4.6 \\
& LTPO
& 58.6 & 45.6 & 12.9 & 9.2 & 21.7 & 29.6 & +7.6 \\
& TTSV
& 68.6 & 46.8 & 20.6 & 8.8 & 24.8 & 33.9 & +11.9 \\

& \cellcolor[gray]{0.94}\textbf{TASCO-RP}
& \cellcolor[gray]{0.94}\underline{70.8}
& \cellcolor[gray]{0.94}\underline{50.6}
& \cellcolor[gray]{0.94}\textbf{22.4}
& \cellcolor[gray]{0.94}\underline{10.4}
& \cellcolor[gray]{0.94}\textbf{27.8}
& \cellcolor[gray]{0.94}\underline{36.4}
& \cellcolor[gray]{0.94}+14.4 \\

& \cellcolor[gray]{0.94}\textbf{TASCO-SAP}
& \cellcolor[gray]{0.94}\textbf{71.8}
& \cellcolor[gray]{0.94}\textbf{52.2}
& \cellcolor[gray]{0.94}\underline{21.3}
& \cellcolor[gray]{0.94}\textbf{12.1}
& \cellcolor[gray]{0.94}\underline{26.8}
& \cellcolor[gray]{0.94}\textbf{36.8}
& \cellcolor[gray]{0.94}+14.8 \\
\midrule

\multirow{7}{*}{Qwen2.5-Math-7B}
& Zero-Shot CoT
& 52.2 & 44.4 & 12.9 & 14.6 & 31.8 & 31.2 & -- \\
& SLOT
& 58.8 & 56.2 & 29.4 & 18.8 & 32.8 & 39.2 & +8.0 \\
& LatentSeek
& 57.6 & 48.1 & 20.5 & 6.2 & 32.8 & 33.0 & +1.8 \\
& LTPO
& 65.6 & 42.8 & 17.6 & 18.8 & 33.5 & 35.7 & +4.5 \\
& TTSV
& 71.0 & 59.1 & 33.5 & 17.9 & 36.4 & 43.6 & +12.4 \\

& \cellcolor[gray]{0.94}\textbf{TASCO-RP}
& \cellcolor[gray]{0.94}\textbf{75.8}
& \cellcolor[gray]{0.94}\textbf{68.8}
& \cellcolor[gray]{0.94}\underline{37.1}
& \cellcolor[gray]{0.94}\underline{21.2}
& \cellcolor[gray]{0.94}\underline{37.4}
& \cellcolor[gray]{0.94}\underline{48.1}
& \cellcolor[gray]{0.94}+16.9 \\

& \cellcolor[gray]{0.94}\textbf{TASCO-SAP}
& \cellcolor[gray]{0.94}\underline{73.4}
& \cellcolor[gray]{0.94}\underline{66.2}
& \cellcolor[gray]{0.94}\textbf{39.1}
& \cellcolor[gray]{0.94}\textbf{24.2}
& \cellcolor[gray]{0.94}\textbf{38.9}
& \cellcolor[gray]{0.94}\textbf{48.4}
& \cellcolor[gray]{0.94}+17.2 \\
\midrule

\multirow{7}{*}{Llama3.1-8B-Instruct}
& Zero-Shot CoT
& 47.4 & 24.7 & 22.1 & 7.5 & 28.3 & 26.0 & -- \\
& SLOT
& 48.8 & 23.8 & 22.8 & 9.2 & \textbf{33.8} & 27.7 & +1.7 \\
& LatentSeek
& \underline{49.4} & 26.8 & 23.2 & 3.3 & 32.3 & 27.0 & +1.0 \\
& LTPO
& 48.4 & 27.5 & 20.5 & \underline{11.3} & 30.5 & 27.6 & +1.6 \\
& TTSV
& 47.8 & 22.8 & 22.1 & 6.2 & 27.8 & 25.3 & -0.7 \\

& \cellcolor[gray]{0.94}\textbf{TASCO-RP}
& \cellcolor[gray]{0.94}46.8
& \cellcolor[gray]{0.94}\textbf{29.1}
& \cellcolor[gray]{0.94}\underline{25.0}
& \cellcolor[gray]{0.94}10.4
& \cellcolor[gray]{0.94}\underline{33.3}
& \cellcolor[gray]{0.94}\underline{28.9}
& \cellcolor[gray]{0.94}+2.9 \\

& \cellcolor[gray]{0.94}\textbf{TASCO-SAP}
& \cellcolor[gray]{0.94}\textbf{50.2}
& \cellcolor[gray]{0.94}\underline{28.4}
& \cellcolor[gray]{0.94}\textbf{26.5}
& \cellcolor[gray]{0.94}\textbf{11.7}
& \cellcolor[gray]{0.94}31.8
& \cellcolor[gray]{0.94}\textbf{29.7}
& \cellcolor[gray]{0.94}+3.7 \\
\bottomrule
\end{NiceTabular}

\caption{
Main performance comparison across three models and five reasoning benchmarks
(accuracy, \%). $\Delta$ denotes the absolute percentage-point change relative
to Zero-Shot CoT. Best results are in \textbf{bold}; second-best results are
\underline{underlined}.
}
\label{tab:main_results}
\end{table*}

\subsection{Experimental Setup}
\paragraph{Models.}
We evaluate TASCO on five open-source LLMs covering both general and reasoning models. General LLMs include Qwen2.5-Math~\citep{yang2024qwen2} in two sizes (1.5B and 7B) and LLaMA3.1-8B-Instruct~\citep{grattafiori2024llama}. Reasoning LLMs include DeepSeek-R1-Distill-Qwen in two sizes (1.5B and 7B)~\citep{guo2025deepseek}.

\paragraph{Datasets.}
We evaluate on six reasoning benchmarks. For mathematics, we use MATH-500~\citep{hendrycks2021measuring}, AMC23, AIME24~\citep{li2024numinamath}, and AIME25~\citep{balunovic2026matharena}, spanning high-school to olympiad-level problems. For science, we use Minerva Math~\citep{lewkowycz2022solving}, with undergraduate-level STEM problems, and GPQA Diamond~\citep{rein2023gpqa}, with graduate-level questions written by domain experts.

\paragraph{Baselines.}
For general LLMs, we include CoT~\citep{wei2022chain} as the standard prompting baseline. We compare with SLOT~\citep{hu2025slot} and LatentSeek~\citep{latentseek}, which perform test-time adaptation through final-layer or latent representations. We also compare with LTPO~\citep{ye2025ltpo} and TTSV~\citep{kang2026model}, which optimize continuous latent variables using model-internal confidence signals while keeping the backbone frozen. For reasoning LLMs, we compare with s1~\citep{muennighoff2025s1}, CoD~\citep{xu2025chain}, and $\alpha1$~\citep{zhang2025alphaone}, which control reasoning behavior at test time. 

\paragraph{Implementation Details.}
For general LLMs, we use a shared prefix of length $L=20$ and a maximum generation length of 3,072 tokens. For reasoning-enhanced LLMs, we use $L=10$ and allow up to 8,192 tokens for longer reasoning traces. We optimize the prefix using AdamW with a batch size of 16. Random Perturbation uses $K=8$ perturbed rollouts with $\sigma=0.01$, while Sharpness-Aware Perturbation uses a perturbation radius of $\rho=0.8$. We further analyze sensitivity to $\sigma$ and $\rho$. All experiments run on a single NVIDIA A800 GPU. More details are provided in Appendix A.

\paragraph{Evaluation.}
We report accuracy on all benchmarks. For mathematical reasoning benchmarks, we extract the final answer and determine correctness using dataset-specific normalization and equivalence rules. For AMC23 and AIME24, we independently sample eight responses for each problem and report the mean accuracy across the 8 generations.

\subsection{Main Results}

Tables~\ref{tab:main_results} and~\ref{tab:reasoning_results} compare TASCO with baselines on general and reasoning-enhanced LLMs, respectively, showing accuracy gains across both model types. Furthermore, steering the model toward high-confidence reasoning states yields shorter outputs, improving token efficiency.

\paragraph{Performance Improvement.}
Tables~\ref{tab:main_results} and~\ref{tab:reasoning_results} show TASCO improves both general and reasoning-enhanced LLMs. On Qwen2.5-Math-1.5B and Qwen2.5-Math-7B, TASCO raises average accuracy from 22.0\% to 36.8\% and from 31.2\% to 48.4\%, gains of 14.8 and 17.2 percentage points over Zero-Shot CoT. TASCO further outperforms TTSV by 2.9 and 4.8 points, respectively. TASCO also improves LLaMA3.1-8B-Instruct from 26.0\% to 29.7\%, demonstrating applicability beyond Qwen. On reasoning-enhanced models, TASCO improves the four-benchmark average from 45.8\% to 52.6\% on DeepSeek-R1-Distill-Qwen-1.5B and from 59.5\% to 65.5\% on DeepSeek-R1-Distill-Qwen-7B, gains of 6.8 and 6.0 percentage points. It surpasses the strongest reasoning-control baseline by 4.3 and 2.5 points, respectively. Overall, incorporating local stability into confidence optimization improves test-time reasoning across model families and reasoning paradigms.

\begin{figure*}[t]
    \centering
    \includegraphics[width=0.98\textwidth]{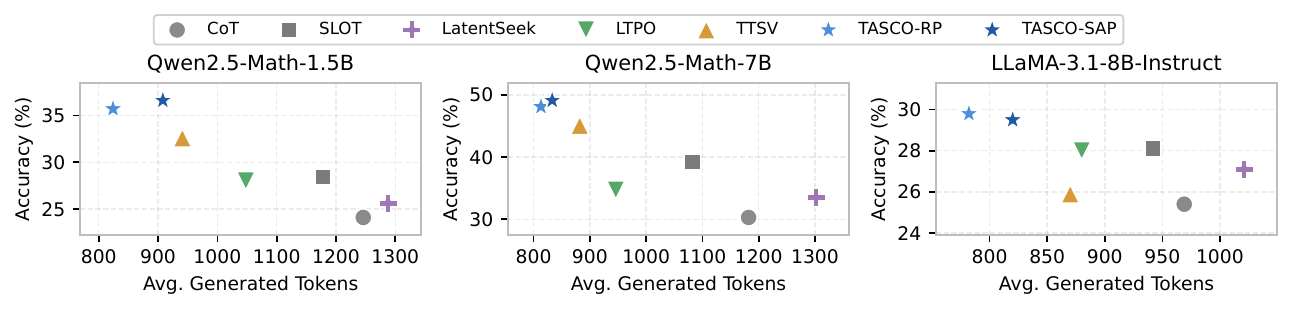}
    \caption{{Accuracy-token trade-off across general LLMs, averaged over all benchmarks.}}
    \label{fig:token_efficiency}
\end{figure*}

\begin{figure*}[t]
    \centering
    \includegraphics[width=0.98\textwidth]{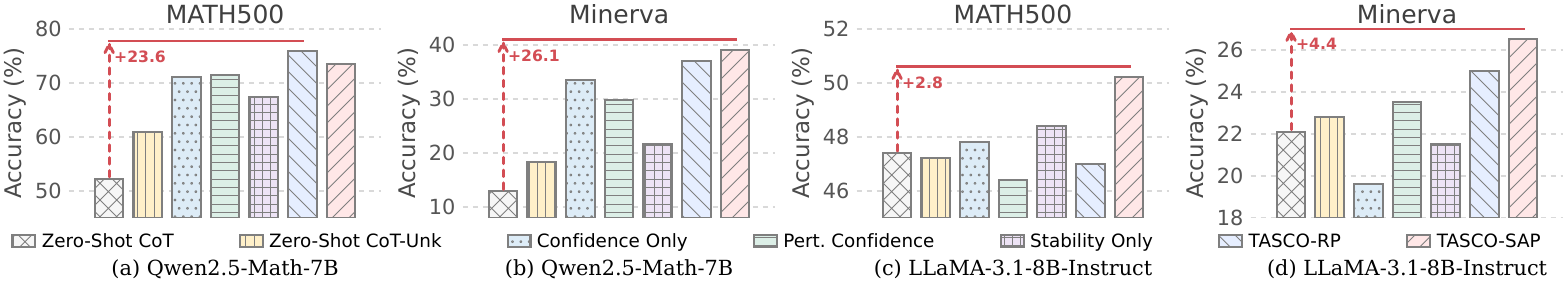}
    \caption{Impact of optimized prefixes on MATH500 and Minerva. The results indicate that the gains of TASCO come from learning stability-aware task-level prefixes rather than simply adding extra prefix positions.}
    \label{fig:prefix_variants}
\end{figure*}

\paragraph{Token Efficiency.}
TASCO also improves token efficiency. Figure~\ref{fig:token_efficiency} shows that TASCO-R and TASCO-S reduce average generation length by 28.1\% and 24.0\%, respectively, across general LLMs compared with CoT. Similar gains on reasoning-enhanced LLMs are reported in Appendix E. These reductions lower inference cost and suggest that TASCO improves accuracy without requiring longer reasoning traces. Instead, the optimized prefix promotes concise, reliable reasoning while improving accuracy.

\begin{table}[t]
\centering
\small
\setlength{\tabcolsep}{3.5pt}
\renewcommand{\arraystretch}{0.9}

\begin{tabularx}{0.98\columnwidth}{
    @{}l*{4}{>{\centering\arraybackslash}X}@{}
}
\toprule
\textbf{Method}
& \textbf{AIME24}
& \textbf{AIME25}
& \textbf{AMC23}
& \textbf{MATH500} \\
\midrule

\multicolumn{5}{c}{\textit{DeepSeek-R1-Distill-Qwen-1.5B}} \\
\midrule
Base & 22.8 & 22.5 & 59.4 & 78.6 \\
s1 & 25.8 & \underline{25.4} & 61.9 & 80.2 \\
CoD & 26.7 & 20.8 & \underline{66.2} & 79.4 \\
$\alpha_1$ & \textbf{30.8} & 12.9 & 63.1 & \underline{81.0} \\
\rowcolor[gray]{0.94}
TASCO & \underline{30.4} & \textbf{29.2} & \textbf{69.4} & \textbf{81.4} \\

\midrule
\multicolumn{5}{c}{\textit{DeepSeek-R1-Distill-Qwen-7B}} \\
\midrule
Base & 44.2 & 25.0 & 79.4 & 89.4 \\
s1 & 37.5 & 24.2 & 78.1 & 89.0 \\
CoD & \underline{45.0} & \underline{33.3} & 85.6 & 88.0 \\
$\alpha_1$ & 36.7 & 32.5 & \textbf{90.6} & \underline{89.8} \\
\rowcolor[gray]{0.94}
TASCO & \textbf{49.2} & \textbf{34.2} & \underline{88.1} & \textbf{90.4} \\
\bottomrule
\end{tabularx}

\caption{Evaluation on reasoning-enhanced LLMs using DeepSeek-R1-Distill models as backbones. TASCO results are obtained with the TASCO-SAP variant.}
\label{tab:reasoning_results}
\end{table}

\subsection{Analyses of TASCO}
\paragraph{Ablation of Prefix Optimization.}

We exam whether TASCO's gains arise from using local stability to improve confidence-guided reasoning. CoT-Unk and Confidence Only test whether additional prefix positions or confidence optimization alone suffice. Perturbed Confidence uses the same $K$ trajectories as TASCO-RP but removes the stability objective, isolating additional rollout computation; Stability Only removes the confidence objective, testing whether stability regularization alone explains the gains. As shown in Fig.~\ref{fig:prefix_variants}, TASCO-RP outperforms both controlled variants. Together with the improvements of both TASCO variants over Confidence Only, these results show that the gains arise neither from additional computation nor from stability regularization in isolation, but from using local stability to guide optimization toward reliable high-confidence reasoning.

\begin{figure}[t]
    \centering
    \includegraphics[width=\columnwidth]{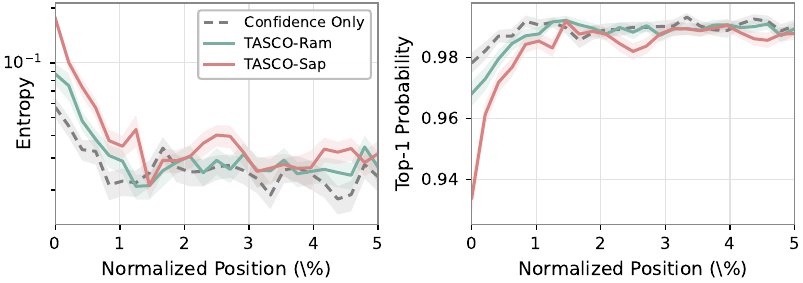}
    \caption{Mean next-token entropy (left) and top-1 probability (right) over the first $5\%$ of decoding steps for Qwen2.5-Math-7B on MATH500.}
    \label{fig:early_dynamics}
\end{figure}

\paragraph{Behavioral Effects of Stability-Aware Optimization.}
We examine how TASCO affects local robustness and confidence formation on Qwen2.5-Math-7B. To assess local robustness, we add Gaussian noise with $\sigma=0.01$ to each learned prefix and generate $K=8$ perturbed rollouts per input. As shown in Table~\ref{tab:local_stability}, confidence-only optimization produces higher trajectory-confidence variance and lower answer consistency across datasets, whereas TASCO reduces this variation and improves consistency. We further analyze confidence formation on MATH500 by tracking the mean next-token entropy and top-1 probability over the first $5\%$ of decoding steps. Figure~\ref{fig:early_dynamics} shows that confidence-only optimization rapidly concentrates probability mass early in generation, whereas TASCO retains higher entropy and lower top-1 probability before the gap narrows later. Together, these results show that TASCO produces reasoning behavior that is more stable under local perturbations and less prone to premature confidence concentration.

\begin{table}[t]
\centering
\small
\setlength{\tabcolsep}{3pt}
\begin{tabular}{@{}lcccc@{}}
\toprule
\textbf{Method}
& \textbf{MATH500}
& \textbf{AMC23}
& \textbf{Minerva}
& \textbf{GPQA} \\
\midrule
\multicolumn{5}{c}{\textit{Confidence Variance ($\times 10^{-4}$) $\downarrow$}} \\
\cmidrule(lr){1-5}
Confidence Only & 4.83 & 3.05 & 6.59 & 15.76 \\
TASCO-RP       & 2.33 & 2.34 & 5.48 & 9.31  \\
TASCO-SAP      & 2.89 & 2.32 & 4.51 & 12.26 \\
\midrule
\multicolumn{5}{c}{\textit{Answer Consistency $\uparrow$}} \\
\cmidrule(lr){1-5}
Confidence Only & 62.9 & 42.5 & 58.9 & 48.1 \\
TASCO-RP       & 71.7 & 47.0 & 62.6 & 59.4 \\
TASCO-SAP      & 73.8 & 50.5 & 66.6 & 52.8 \\
\bottomrule
\end{tabular}
\caption{Local stability under prefix perturbations on Qwen2.5-Math-7B. Lower variance and higher consistency indicate more stable reasoning behavior.}
\label{tab:local_stability}
\end{table}

\begin{figure*}[t]
    \centering
    \includegraphics[width=\textwidth]{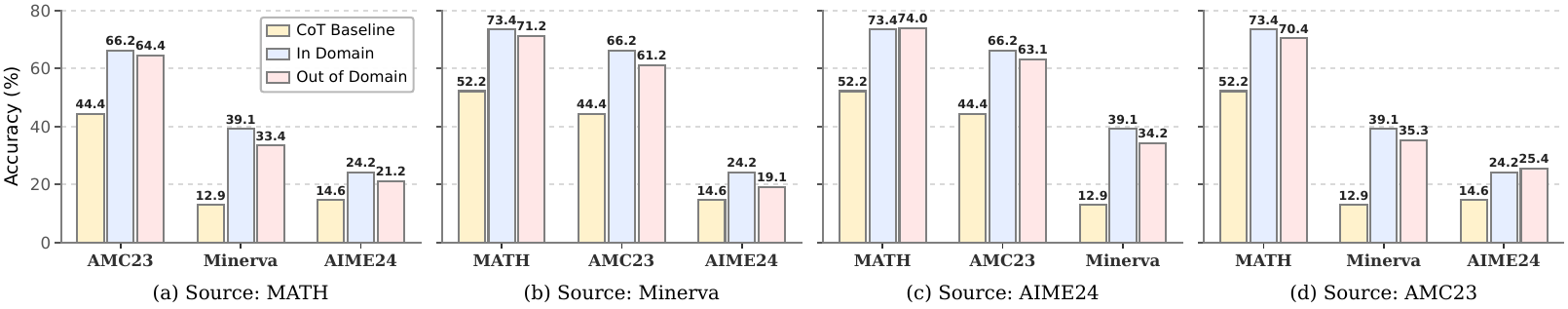}
    \caption{Cross-distribution generalization of TASCO on Qwen2.5-Math-7B, measured by accuracy gains over Zero-Shot CoT.}
    \label{fig:distribution}
\end{figure*}

\paragraph{Analysis of Cross-Distribution Generalization.}
To assess the transferability of the guidance learned by TASCO, we optimize a prefix on one source benchmark and directly apply it to target benchmarks without re-optimization. As shown in Fig.~\ref{fig:distribution}, the transferred prefixes consistently outperform Zero-Shot CoT across all evaluated source-target pairs, indicating that the learned guidance is not confined to its optimization distribution. This transferability stems from TASCO's stability-aware objective: by promoting confidence that remains stable under local perturbations, TASCO guides the model to learn reliable reasoning patterns shared across mathematical tasks rather than merely increasing confidence on the source distribution.

\paragraph{Performance Across Query Difficulty.}
We first estimate query difficulty from eight responses sampled by the unadapted model: queries answered correctly 0--2, 3--5, and 6--8 times are categorized as hard, medium, and easy, respectively. The resulting groups contain 195, 130, and 175 queries for Qwen2.5-Math-7B and 216, 92, and 192 queries for LLaMA3.1-8B-Instruct. We then compare TASCO with confidence-only optimization within the same groups. As shown in Table~\ref{tab:difficulty_performance}, TASCO yields clear gains on hard and medium queries across both models while maintaining comparable performance on easy queries. These results suggest that TASCO does more than reinforce solutions the model already finds easy; it learns reasoning guidance that supports more reliable reasoning on challenging problems.

\begin{table}[t]
\centering
\small
\setlength{\tabcolsep}{5pt}
\begin{tabular}{lcccc}
\toprule
& \multicolumn{2}{c}{Qwen2.5-Math-7B}
& \multicolumn{2}{c}{LLaMA3.1-8B} \\
\cmidrule(lr){2-3}\cmidrule(lr){4-5}
Difficulty & Confidence & TASCO & Confidence & TASCO \\
\midrule
Hard    & 46.7 & 54.4 & 9.7 & 14.4 \\
Medium  & 75.4 & 80.8 & 48.9 & 54.3 \\
Easy    & 94.9 & 96.0 & 90.1 & 88.5 \\
\midrule
Overall & 71.0 & 75.8 & 47.8 & 50.2 \\
\bottomrule
\end{tabular}
\caption{Accuracy (\%) of confidence-only optimization and TASCO-RP across MATH500 difficulty groups.}
\label{tab:difficulty_performance}
\end{table}

\begin{figure}[t]
    \centering
    \includegraphics[width=\columnwidth]{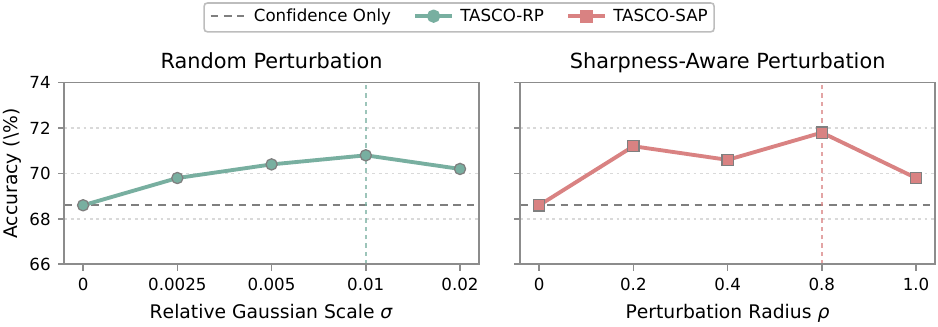}
    \caption{Sensitivity of TASCO to the perturbation scale on Qwen2.5-Math-1.5B with MATH500.}
    \label{fig:scale_sensitivity}
\end{figure}

\paragraph{Sensitivity to Perturbation Scale.}
We evaluate sensitivity to the perturbation scale on Qwen2.5-Math-1.5B with MATH500. For Random Perturbation, we vary the Gaussian scale $\sigma$; for Sharpness-Aware Perturbation, we vary the radius $\rho$, while fixing all other settings. Figure~\ref{fig:scale_sensitivity} shows that both variants outperform confidence-only optimization across a broad range of nonzero scales. Accuracy peaks near the selected defaults and decreases only when the perturbation becomes excessively large, indicating that TASCO does not rely on narrow hyperparameter tuning. Additional sensitivity analyses are provided in Appendix E.

\section{Conclusion}

We observed that high-confidence reasoning is more likely to yield correct answers when it remains stable under local perturbations. Based on this observation, we proposed TASCO, a label-free framework that optimizes a shared task-level prefix for confidence and stability while keeping the LLM frozen. Random Perturbation captures distributional variation across nearby behaviors, whereas Sharpness-Aware Perturbation targets worst-case local sensitivity. Experiments across diverse LLMs and reasoning benchmarks show that stability-aware confidence optimization improves reasoning accuracy and generation efficiency without parameter updates or external supervision.

\bibliography{aaai2027}

\clearpage
\appendix
\section{Appendix}

\section{A. Experimental Details}
\label{app:implementation_details}

This section provides the implementation details of TASCO and the compared baselines. We first describe the prefix optimization and perturbation procedures, followed by the baseline implementations and evaluation protocol.

\begin{table*}[t]
\centering
\small
\setlength{\tabcolsep}{5pt}
\begin{tabular}{lcccccccc}
\toprule
Model family & Prefix length & Batch size & Learning rate & Temp. & Top-$p$ & Rep. penalty & Conf. tokens & Max. tokens \\
\midrule
Qwen2.5-Math & 20 & 16 & 1e-3 & 0.7 & 0.95 & 1.15 & 256 & 3,072 \\
LLaMA3.1-Instruct & 20 & 16 & 5e-6 & 0.7 & 0.95 & 1.05 & 256 & 3,072 \\
DeepSeek-R1-Distill-Qwen & 10 & 16 & 5e-4 & 0.6 & 0.95 & 1.00 & 1,024 & 8,192 \\
\bottomrule
\end{tabular}
\caption{Model-specific optimization and generation settings. ``Conf. tokens'' denotes the maximum number of generated tokens used to compute the confidence objective.}
\label{tab:optimization_settings}
\end{table*}

\paragraph{Optimization and Generation Settings.}
We optimize only the continuous prefix $V$ using AdamW while keeping all parameters of $M_\theta$ frozen. During optimization, trajectories are generated using the model-specific temperature, top-$p$, and repetition penalty reported in Table~\ref{tab:optimization_settings}. The confidence objective is computed from at most the first $T_{\mathrm{opt}}$ generated tokens, reported as ``Conf. tokens.'' Final evaluation uses greedy decoding subject to the corresponding maximum generation length.

\paragraph{Prefix Initialization.}
TASCO can be optimized directly from a freshly initialized prefix. We use model-family-specific initialization. Prefixes for Qwen-based models, including DeepSeek-R1-Distill-Qwen, are sampled from $\mathcal{N}(0,I)$. For LLaMA models, we first encode the unlabeled optimization set and compute the per-dimension empirical mean and variance of the resulting token embeddings. Each prefix vector is then sampled from the Gaussian distribution parameterized by these statistics. Since perturbation-based stability optimization requires additional rollouts, applying it before the prefix has been guided toward higher-confidence reasoning may be inefficient. In addition to direct optimization from scratch, we also support an initial confidence-only stage of 10 epochs, followed by five epochs of TASCO optimization. All stages use the same learning rate reported in Table~\ref{tab:optimization_settings}.

\paragraph{Perturbation Settings.}
For Random Perturbation, we draw $K=8$ Gaussian perturbations with $\sigma=0.01$ at each optimization step. We set the stability weight to $\lambda_{\mathrm{rand}}=20$ for Qwen2.5-Math and $\lambda_{\mathrm{rand}}=5$ for LLaMA3.1-Instruct and DeepSeek-R1-Distill-Qwen. For Sharpness-Aware Perturbation, we use a perturbation radius of $\rho=0.8$. The two variants are optimized and evaluated separately.

\paragraph{Trajectory Generation and Updates.}
At each optimization step, the model generates one trajectory for every input in the current mini-batch using the current prefix. The generated tokens are held fixed during backpropagation, and gradients are computed only with respect to $V$. New trajectories are generated after each prefix update. Random Perturbation independently decodes a trajectory under each perturbed prefix, whereas SAP evaluates its gradient-guided perturbation using the trajectory fixed within the current update.

\subsection{Baseline Implementation Details}

We use the official implementations and follow the configurations reported by the original authors whenever available.

\paragraph{Chain-of-Thought.}
CoT elicits explicit step-by-step reasoning through natural-language instructions without modifying the model.

\paragraph{SLOT.}
SLOT adapts each test query by optimizing a sample-specific vector added to the model's final hidden representations.

\paragraph{LatentSeek.}
LatentSeek uses policy gradients and self-generated rewards to iteratively optimize instance-specific latent reasoning representations.

\paragraph{LTPO.}
LTPO optimizes latent thought vectors at test time using an intrinsic confidence reward while keeping the LLM frozen.

\paragraph{TTSV.}
TTSV learns a task-level continuous prefix by minimizing predictive entropy over the unlabeled test set.

\paragraph{s1.}
s1 applies budget forcing to extend reasoning by inserting additional \texttt{wait} tokens before the model terminates its thinking process.

\paragraph{Chain of Draft.}
CoD encourages concise reasoning by prompting the model to express each intermediate step using only a few words.

\paragraph{$\alpha1$.}
$\alpha1$ controls the transition from slow to fast reasoning through a dynamically scheduled $\alpha$-moment.

\paragraph{Prompt Design.}
We use fixed model-family-specific prompt templates throughout optimization and evaluation. Table~\ref{tab:case_aime24} provides a complete example of the \texttt{qwen25-math-cot} prompt, the embedding-level prefix placement, and the resulting response format. The remaining templates are included in the released implementation.

\paragraph{Evaluation.}
For MATH500, Minerva Math, and GPQA, the main tables report results from a single run. For AMC23 and AIME24, whose smaller test sets produce greater accuracy fluctuations, we independently run each experiment with eight random seeds and report the mean accuracy.

\section{B. Exploring Local Confidence Stability}
\label{app:perturbation_analysis}

We investigate whether local confidence stability is associated with answer correctness and whether this relationship persists after accounting for confidence magnitude and query difficulty. We first establish the basic association and then examine these two alternative explanations.

\subsection{Experimental Setup.}
We conduct the analysis on the full MATH500 benchmark using five models from three families: Qwen2.5-Math-1.5B and 7B, LLaMA3.1-8B-Instruct, and DeepSeek-R1-Distill-Qwen-1.5B and 7B. For each model, we use the final task-level prefix $V$ obtained through confidence-only optimization. For every query $X_i$, we sample $K=8$ Gaussian perturbations with $\sigma=0.01$:
\begin{equation}
V^{(k)}=V+\epsilon^{(k)},\qquad
\epsilon^{(k)}\sim\mathcal{N}(0,\sigma^2I).
\label{eq:analysis_perturbation}
\end{equation}
Under each perturbed prefix, the model greedily decodes a trajectory $Y_i^{(k)}$. Its trajectory-level confidence is defined as the length-normalized log-probability:
\begin{equation}
c_i^{(k)}=\frac{1}{T_i^{(k)}} \sum_{t=1}^{T_i^{(k)}}
\log p_\theta\left(y_{i,t}^{(k)} \mid X_i,V^{(k)},y_{i,<t}^{(k)} \right),
\label{eq:analysis_trajectory_confidence}
\end{equation}
where $y_{i,t}^{(k)}$ denotes the $t$-th token in $Y_i^{(k)}$ and $T_i^{(k)}$ is the total sequence length. We quantify local confidence stability using the sample variance:
\begin{equation}
\mathcal{S}_i=\frac{1}{K-1}\sum_{k=1}^{K}\left(c_i^{(k)}-\bar c_i\right)^2, \qquad
\bar c_i=\frac{1}{K}\sum_{k=1}^{K}c_i^{(k)},
\label{eq:analysis_local_variance}
\end{equation}
where lower $\mathcal{S}_i$ indicates greater local stability, and $\bar{c}_i$ denotes the mean confidence across the perturbed trajectories.

For each query, we additionally greedily decode a reference trajectory under the unperturbed prefix $V$. We evaluate its final answer correctness using the standard benchmark evaluator and compute its original trajectory-level confidence $c_i^{(0)}$ following Eq.~\ref{eq:analysis_trajectory_confidence}.

\subsection{Accuracy across Stability Groups}

\begin{figure*}[t]
    \centering
    \includegraphics[width=0.95\textwidth]{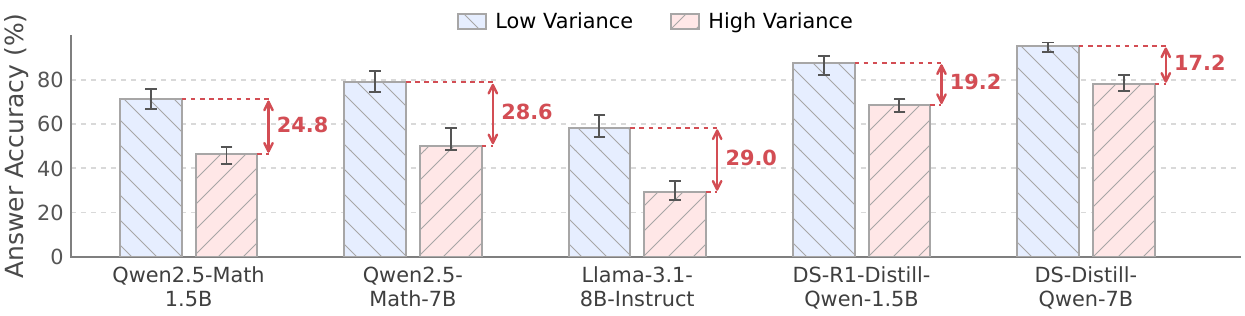}
    \caption{Answer accuracy under the unperturbed prefix for low- and high-variance query groups across five models. Groups are defined by the model-specific median of $\mathcal{S}_i$; error bars show 95\% bootstrap confidence intervals, and red annotations indicate accuracy gaps.}
    \label{fig:stability_correctness}
\end{figure*}

We first examine whether local confidence stability is associated with
correctness at the unperturbed prefix. For each query, we estimate
$\mathcal{S}_i$ from perturbed trajectories and independently evaluate
correctness using a trajectory decoded under the unperturbed prefix. We split
queries at the model-specific median of $\mathcal{S}_i$ and report group
accuracy with 95\% bootstrap confidence intervals.

\begin{table}[t]
\centering
\small
\setlength{\tabcolsep}{5pt}
\begin{tabular}{lcc}
\toprule
Model
& Spearman's $\rho$
& AUROC \\
\midrule
Qwen2.5-Math-1.5B & -0.369 & 0.668 \\
Qwen2.5-Math-7B & -0.403 & 0.687 \\ 
LLaMA3.1-8B-Instruct & -0.413 & 0.683 \\ 
DS-Distill-Qwen-1.5B & -0.308 & 0.709 \\
DS-Distill-Qwen-7B & -0.359 & 0.745 \\
\bottomrule
\end{tabular}
\caption{Association between local confidence variance and unperturbed correctness. Spearman's $\rho$ measures the correlation between $\mathcal{S}_i$ and correctness, while AUROC uses $-\mathcal{S}_i$ to distinguish correct from incorrect answers.}
\label{tab:stability_center_correctness}
\end{table}

As shown in Figure~\ref{fig:stability_correctness}, Low-Variance queries achieve
higher unperturbed accuracy across all five models, with gaps ranging from 17.2
to 29.0 percentage points. This pattern holds for both general and
reasoning-enhanced LLMs. Table~\ref{tab:stability_center_correctness} further
shows consistently negative correlations between local variance and
unperturbed correctness, with AUROCs ranging from 0.668 to 0.745. Importantly,
stability is estimated from perturbed trajectories, whereas correctness is
evaluated independently at the unperturbed prefix. The observed separation is
therefore not merely induced by measuring variance and accuracy on the same
rollouts. Instead, it suggests that neighborhood stability contains information
about the reliability of the optimized prefix itself across model families.

\subsection{Controlling for Confidence Magnitude.}

\begin{table}[t]
\centering
\small
\setlength{\tabcolsep}{3.5pt}
\begin{tabular}{lcccc}
\toprule
Model
& Gap
& $p_{\mathrm{perm}}$
& $\rho_{\mathrm{partial}}$
& $p_{\mathrm{partial}}$ \\
\midrule
Qwen2.5-Math-1.5B & 5.4 & 0.086 & -0.108 & 0.128 \\
Qwen2.5-Math-7B & 11.6 & <0.001 & -0.242 & <0.001 \\
LLaMA3.1-8B & 23.3 & <0.001 & -0.283 & <0.001 \\
DS-Distill-Qwen-1.5B & 30.5 & <0.001 & -0.560 & <0.001 \\
DS-Distill-Qwen-7B & 8.7 & <0.001 & -0.297 & <0.001 \\
\bottomrule
\end{tabular}
\caption{Confidence-controlled association between local variance and
perturbation accuracy. Gap$_{\mathrm{strat}}$ is the Low- minus High-Variance
accuracy after five-way stratification by mean confidence. Partial Spearman
correlation controls for mean confidence continuously.}
\label{tab:stability_controlled}
\end{table}

A remaining concern is that local variance may merely reflect confidence magnitude, as more confident queries may also be inherently more stable. To separate these effects, we divide queries into five equal-frequency strata according to their mean confidence across perturbed trajectories. Within each stratum, we split queries at the median of $\mathcal{S}_i$ and compare their perturbation accuracy, thereby reducing confidence differences between the Low- and High-Variance groups. We estimate confidence intervals using stratified bootstrap sampling and assess the accuracy gap with a stratified permutation test. We additionally compute the partial Spearman correlation between local variance and perturbation accuracy while controlling for mean confidence.

As shown in Table~\ref{tab:stability_controlled}, Low-Variance queries achieve higher perturbation accuracy after confidence stratification across all five models. The differences are statistically significant on four models, with accuracy gaps ranging from 11.6 to 30.5 percentage points. Partial Spearman analysis yields the same pattern, showing significant negative correlations after controlling for mean confidence on these four models. These results suggest that local stability generally provides reliability information beyond confidence magnitude.

Qwen2.5-Math-1.5B is an exception, showing a non-significant 5.4-point gap and a near-zero partial correlation ($\rho=-0.108$, $p=0.128$). Its local variance is strongly correlated with mean confidence ($\rho=-0.717$), suggesting that variance provides little additional information beyond confidence magnitude for this model.

Overall, these analyses show a consistent association between local confidence stability and answer accuracy across model families. The relationship persists among queries with comparable mean confidence and within different difficulty groups, indicating that it cannot be explained solely by confidence magnitude or easier queries. These findings provide empirical motivation for incorporating local stability into confidence-based test-time optimization. The main experiments further examine whether optimizing this signal improves reasoning performance.

\section{C. Proofs and Additional Theoretical Analysis}

The main paper presents the local-sensitivity results for Random Perturbation and Sharpness-Aware Perturbation. This section restates the quantities required by the proofs, makes the assumptions explicit, and provides the complete derivations. We distinguish behavioral confidence obtained after re-decoding from smooth confidence evaluated on a fixed trajectory, separating variation across decoded trajectories from local variation within a fixed decoding region.

\subsection{Setup and Assumptions}

We identify the prefix $V\in\mathbb{R}^{L\times d}$ with a vector in
$\mathbb{R}^{m}$, where $m=Ld$. Under this vectorization, the Euclidean norm and inner product coincide with the Frobenius norm and inner product. For Random Perturbation, we draw $K$ independent Gaussian perturbations:
\begin{equation}
V^{(k)}=V+\epsilon^{(k)}, \qquad
\epsilon^{(k)}
\sim
\mathcal{N}(0,\sigma^2I).
\label{eq:app_random_prefix}
\end{equation}
Under each perturbed prefix $V^{(k)}$, the model greedily decodes a trajectory
$Y^{(k)}=(y_1^{(k)},\ldots,y_{T_k}^{(k)})$. Following the notation in the main paper, its trajectory-level confidence is:
\begin{equation}
c^{(k)} = \frac{1}{T_k}
\sum_{t=1}^{T_k}
\log p_\theta
\left( y_t^{(k)} \mid E'^{(k)}(X),y_{<t}^{(k)}
\right),
\label{eq:trajectory_confidence}
\end{equation}
where $E'^{(k)}(X)=[V^{(k)};E(X)]$ denotes the input conditioned on
$V^{(k)}$. Let
$\bar c=K^{-1}\sum_{k=1}^{K}c^{(k)}$. The Random Perturbation objective uses the sample variance:
\begin{equation}
\mathcal{S}(V;X)=\frac{1}{K-1}
\sum_{k=1}^{K}
\left(c^{(k)}-\bar c\right)^2.
\label{eq:random_sharpness}
\end{equation}

For the population analysis, draw
$\epsilon\sim\mathcal{N}(0,\sigma^2I)$.
Let $Z$ denote the greedy trajectory under $V+\epsilon$, and let
$C=c(V+\epsilon)$ denote its trajectory-level confidence, defined analogously to Eq.~\ref{eq:trajectory_confidence}. Thus, each $c^{(k)}$ is an independent realization of $C$. We assume that
$\mathbb{E}[C^2]<\infty$.

For the smooth within-trajectory analysis, fix a trajectory
$Y=(y_1,\ldots,y_T)$ and define its teacher-forced confidence as:
\begin{equation}
f_Y(V) = \frac{1}{T}
\sum_{t=1}^{T} \log p_\theta \left( y_t \mid E'(X),y_{<t} \right),
\label{eq:fixed_trajectory_confidence}
\end{equation}
where $E'(X)=[V;E(X)]$ is the prefix-conditioned input.

We assume that $V$ is an interior point of the decoding region associated with $Y$: there exists $r>0$ such that every prefix $V+\delta$ satisfying
$\|\delta\|_F\leq r$ greedily decodes $Y$. Let
$g_Y=\nabla_V f_Y(V)$ and
$H_Y=\nabla_V^2 f_Y(V)$.

We further assume that $f_Y$ is sufficiently smooth around $V$ so that, for
$U\sim\mathcal{N}(0,I)$:
\begin{equation}
\begin{aligned}
f_Y(V+\sigma U)
={}&f_Y(V)+\sigma\langle g_Y,U\rangle_F \\
&+\frac{\sigma^2}{2}
\langle U,H_YU\rangle_F+R_\sigma(U),
\end{aligned}
\label{eq:app_random_taylor}
\end{equation}
where
$\mathbb{E}[R_\sigma(U)^2]=\mathcal{O}(\sigma^6)$.
This condition follows, for example, from a locally bounded third derivative together with an integrable Taylor remainder. As $\sigma\to0$, the probability that a Gaussian perturbation leaves the fixed-decoding neighborhood decreases faster than any polynomial in $\sigma$. Under the same integrability condition, these tail events do not affect the polynomial orders derived below.

For Sharpness-Aware Perturbation, fix a trajectory $Y$ within an update and let
$\ell(V)=\mathcal{L}_{\mathrm{conf}}(V;X,Y)$ denote the corresponding fixed-trajectory confidence loss. We assume that $\ell$ has a $\beta$-Lipschitz-continuous gradient on
$\{V+\epsilon:\|\epsilon\|_F\leq\rho\}$. The stationary-point result additionally assumes that $\ell$ is twice continuously differentiable in a neighborhood of $V$.

\subsection{Random Perturbation}

\paragraph{Proposition 1 (Population behavioral sharpness).}
Let $C_1,\ldots,C_K$ be the trajectory-level confidences induced by $K$ independent Gaussian perturbations, and let $\mathcal{S}(V;X)$ be their sample variance in Eq.~\ref{eq:random_sharpness}. Then:
\begin{equation}
\mathbb{E}\!\left[\mathcal{S}(V;X)\right]
=\mathcal{S}_\sigma(V;X)=\operatorname{Var}_{\epsilon\sim\mathcal{N}(0,\sigma^2I)}[C].
\label{eq:app_unbiased_sharpness}
\end{equation}
Moreover, the population sharpness satisfies:
\begin{equation}
\mathcal{S}_\sigma(V;X)=\mathbb{E}_{Z}\!\left[\operatorname{Var}(C\mid Z)\right]+\operatorname{Var}_{Z}\!\left(\mathbb{E}[C\mid Z]\right).
\label{eq:app_total_variance}
\end{equation}

\paragraph{Proof.}
Let $\mu=\mathbb{E}[C]$ and $\bar{C}=K^{-1}\sum_{k=1}^{K}C_k$. The following identity holds:
\begin{equation}
\sum_{k=1}^{K}(C_k-\bar{C})^2=\sum_{k=1}^{K}(C_k-\mu)^2-
K(\bar{C}-\mu)^2.
\label{eq:app_sample_variance_identity}
\end{equation}
Taking expectations gives:
\begin{equation}
\begin{aligned}
\mathbb{E}\!\left[
\sum_{k=1}^{K}(C_k-\bar{C})^2
\right]
&=K\operatorname{Var}(C)-K\operatorname{Var}(\bar{C})\\&=(K-1)\operatorname{Var}(C),
\end{aligned}
\end{equation}
where independence implies
$\operatorname{Var}(\bar{C})=\operatorname{Var}(C)/K$. Dividing by
$K-1$ proves Eq.~\ref{eq:app_unbiased_sharpness}.

For the second result, define $m(Z)=\mathbb{E}[C\mid Z]$. The centered confidence decomposes as:
\begin{equation}
C-\mathbb{E}[C]=\bigl(C-m(Z)\bigr)+\bigl(m(Z)-\mathbb{E}[C]\bigr).
\end{equation}
The cross term has zero expectation because
$\mathbb{E}[C-m(Z)\mid Z]=0$. Taking the expectation of the squared identity yields Eq.~\ref{eq:app_total_variance}. This argument treats $Z$ as a discrete random variable and does not differentiate through decoding.
\hfill$\square$

\paragraph{Corollary 1 (Control of large confidence deviations).}
For any $\tau>0$, Chebyshev's inequality gives:
\begin{equation}
\Pr\!\left(
\left|C-\mathbb{E}[C]\right|\geq\tau
\right)
\leq
\frac{\mathcal{S}_\sigma(V;X)}{\tau^2}.
\label{eq:app_random_concentration}
\end{equation}
Thus, reducing population behavioral sharpness controls the probability of large confidence deviations under the perturbation distribution. This is a stability statement and does not by itself imply that the decoded answer is correct.

\paragraph{Corollary 2 (Contribution of behavioral changes).}
Let $Y_0$ be the trajectory decoded at the unperturbed prefix, and let $\phi(Z)$ denote a discrete behavioral descriptor of trajectory $Z$, such as its normalized final answer. Define $B=\mathbf{1}\{\phi(Z)\neq\phi(Y_0)\}$ and $p_{\mathrm{chg}}=\Pr(B=1)$. For $0<p_{\mathrm{chg}}<1$, define the conditional means:
\begin{equation}
\mu_{\mathrm{same}}=\mathbb{E}[C\mid B=0],\qquad
\mu_{\mathrm{chg}}=\mathbb{E}[C\mid B=1].
\end{equation}
Because $B$ is a function of $Z$, applying the law of total variance to
$m(Z)=\mathbb{E}[C\mid Z]$ gives:
\begin{equation}
\operatorname{Var}_{Z}\!\left(\mathbb{E}[C\mid Z]\right) \geq p_{\mathrm{chg}}(1-p_{\mathrm{chg}})
\left(\mu_{\mathrm{chg}}-\mu_{\mathrm{same}}\right)^2.
\label{eq:app_behavior_change_lower_bound}
\end{equation}

The right-hand side is $\operatorname{Var}_{B}\!\left(\mathbb{E}[C\mid B]\right)$, which measures the confidence difference between the unchanged and changed groups. Thus, the bound is informative when behavioral changes are accompanied by a shift in mean confidence. If the two groups have identical mean confidence, it provides no information about how frequently behavior changes.

\paragraph{Proposition 2 (Smooth local sensitivity).}
Within a neighborhood where greedy decoding remains $Y$, the local confidence variance satisfies the following expansion as $\sigma\to0$:
\begin{equation}
\operatorname{Var}_{\epsilon}[f_Y(V+\epsilon)]
=
\sigma^2\|\nabla_Vf_Y(V)\|_F^2
+
\mathcal{O}(\sigma^4).
\label{eq:app_random_first_order}
\end{equation}
If $\nabla_Vf_Y(V)=0$, the corresponding expansion is:
\begin{equation}
\operatorname{Var}_{\epsilon}[f_Y(V+\epsilon)]
=
\frac{\sigma^4}{2}
\|\nabla_V^2f_Y(V)\|_F^2
+
o(\sigma^4).
\label{eq:app_random_second_order}
\end{equation}

\paragraph{Proof.}
Write $\epsilon=\sigma U$, where $U\sim\mathcal{N}(0,I)$, and abbreviate $g=g_Y$ and $H=H_Y$. From Eq.~\ref{eq:app_random_taylor}, define:
\begin{equation}
A=\langle g,U\rangle_F,\qquad
Q=\frac{1}{2}\langle U,HU\rangle_F.
\end{equation}
Then,
$f_Y(V+\sigma U)=f_Y(V)+\sigma A+\sigma^2Q+R_\sigma(U)$.
The Gaussian moments satisfy:
\begin{equation}
\mathbb{E}[A]=0,\qquad
\operatorname{Var}(A)=\|g\|_F^2,\qquad
\operatorname{Cov}(A,Q)=0.
\end{equation}
The last equality follows because every third-order centered Gaussian moment is zero. Moreover, $\operatorname{Var}(\sigma^2Q)=\mathcal{O}(\sigma^4)$. By the Cauchy--Schwarz inequality and $\mathbb{E}[R_\sigma(U)^2]=\mathcal{O}(\sigma^6)$, the remainder and its covariance terms contribute at most $\mathcal{O}(\sigma^4)$. Consequently, we obtain:
\begin{equation}
\operatorname{Var}[f_Y(V+\sigma U)]
=
\sigma^2\|g\|_F^2+\mathcal{O}(\sigma^4).
\end{equation}
This proves Eq.~\ref{eq:app_random_first_order}.

At a stationary point of $f_Y$, $g=0$, and the quadratic term becomes dominant:
\begin{equation}
\operatorname{Var}[f_Y(V+\sigma U)]
=
\sigma^4\operatorname{Var}(Q)+o(\sigma^4).
\end{equation}
For a symmetric Hessian $H$ and a standard Gaussian vector $U$, the following identities hold:
\begin{equation}
\mathbb{E}[U^\top HU]=\operatorname{tr}(H),\qquad
\operatorname{Var}(U^\top HU)=2\|H\|_F^2.
\end{equation}
Therefore,
$\operatorname{Var}(Q)=\frac{1}{2}\|H\|_F^2$, which proves
Eq.~\ref{eq:app_random_second_order}.
\hfill$\square$

This proposition characterizes the smooth within-trajectory component of behavioral sharpness; variation due to trajectory switching is captured by the between-trajectory term in Eq.~\ref{eq:app_total_variance}.

\paragraph{Perturbation scale.}
The elementwise standard deviation $\sigma$ corresponds to the following root-mean-square Frobenius radius:
\begin{equation}
\sqrt{\mathbb{E}\!\left[\|\epsilon\|_F^2\right]}
=
\sigma\sqrt{Ld}.
\label{eq:app_random_radius}
\end{equation}
Thus, choosing $\sigma=\rho/\sqrt{Ld}$ would match the root-mean-square norm of the random perturbation to the SAP radius $\rho$. This relation aligns only their geometric scales and does not make the two objectives equivalent.

\subsection{Sharpness-Aware Perturbation}

\paragraph{Proposition 3 (First-order worst-case bound).}
Let the worst-case local increase be:
\begin{equation}
\Delta_\rho(V)=\max_{\|\epsilon\|_F\leq\rho}
\left[\ell(V+\epsilon)-\ell(V)\right].
\end{equation}
Let $g=\nabla_V\ell(V)$. If $\ell$ has a
$\beta$-Lipschitz-continuous gradient on the perturbation ball, the following bound holds:
\begin{equation}
\left|
\Delta_\rho(V)-\rho\|g\|_F
\right|
\leq
\frac{\beta\rho^2}{2}.
\label{eq:app_sap_worst_case}
\end{equation}
For $g\neq0$, the gradient-guided perturbation
$\epsilon^\star=\rho g/\|g\|_F$ additionally satisfies:
\begin{equation}
\left|
\ell(V+\epsilon^\star)-\ell(V)-\rho\|g\|_F
\right|
\leq
\frac{\beta\rho^2}{2}.
\label{eq:app_sap_perturbed_loss}
\end{equation}
Accordingly, the implemented single-example SAP loss is
$\mathcal{L}_{\mathrm{SAP}}=\ell(V+\epsilon^\star)$.

\paragraph{Proof.}
For any $\epsilon$ in the perturbation ball, the fundamental theorem of calculus gives:
\begin{equation}
\begin{aligned}
&\ell(V+\epsilon)-\ell(V)-\langle g,\epsilon\rangle_F \\
&\qquad=
\int_0^1
\left\langle
\nabla_V\ell(V+t\epsilon)-g,\epsilon
\right\rangle_F\,dt.
\end{aligned}
\end{equation}
Gradient Lipschitzness and Cauchy--Schwarz imply:
\begin{equation}
\left|
\ell(V+\epsilon)-\ell(V)-\langle g,\epsilon\rangle_F
\right| \leq\int_0^1\beta t\|\epsilon\|_F^2\,dt=
\frac{\beta}{2}\|\epsilon\|_F^2.
\label{eq:app_descent_lemma}
\end{equation}
For the upper bound, Eq.~\ref{eq:app_descent_lemma} gives:
\begin{equation}
\Delta_\rho(V)
\leq\max_{\|\epsilon\|_F\leq\rho}
\langle g,\epsilon\rangle_F+\frac{\beta\rho^2}{2}=\rho\|g\|_F+\frac{\beta\rho^2}{2}.
\end{equation}
When $g\neq0$, evaluating the inner objective at
$\epsilon^\star=\rho g/\|g\|_F$ gives:
\begin{equation}
\Delta_\rho(V)
\geq
\ell(V+\epsilon^\star)-\ell(V)
\geq
\rho\|g\|_F-\frac{\beta\rho^2}{2}.
\end{equation}
When $g=0$, choosing $\epsilon=0$ gives
$\Delta_\rho(V)\geq0$, while the upper bound gives
$\Delta_\rho(V)\leq\beta\rho^2/2$. Hence,
Eq.~\ref{eq:app_sap_worst_case} holds in both cases. Finally, substituting
$\epsilon^\star$ into Eq.~\ref{eq:app_descent_lemma} proves
Eq.~\ref{eq:app_sap_perturbed_loss}.
\hfill$\square$

\paragraph{Effect of stop-gradient.}
For $g\neq0$, the bounds above concern the value of the perturbed objective and are unchanged by the stop-gradient operation. Treating $\epsilon^\star$ as constant gives the outer gradient:
\begin{equation}
\nabla_V^{\mathrm{sg}}\mathcal{L}_{\mathrm{SAP}}=\nabla_V\ell(V+\epsilon^\star).
\label{eq:app_stop_gradient}
\end{equation}
Gradient Lipschitzness further gives:
\begin{equation}
\left\|
\nabla_V^{\mathrm{sg}}\mathcal{L}_{\mathrm{SAP}}-g
\right\|_F
\leq
\beta\rho.
\label{eq:app_stop_gradient_bound}
\end{equation}
Thus, stop-gradient removes derivatives through the construction of
$\epsilon^\star$ while retaining the gradient evaluated at the perturbed prefix.

\paragraph{Proposition 4 (Stationary-point curvature).}
Suppose that $\nabla_V\ell(V)=0$ and that $\ell$ is twice continuously differentiable in a neighborhood of $V$. As $\rho\to0$, the ideal inner maximization satisfies:
\begin{equation}
\Delta_\rho(V)=\frac{\rho^2}{2}
\left[
\lambda_{\max}\!\left(\nabla_V^2\ell(V)\right)
\right]_+ +o(\rho^2),
\label{eq:app_sap_second_order}
\end{equation}
where $[a]_+=\max(a,0)$.

\paragraph{Proof.}
Let $H=\nabla_V^2\ell(V)$. Twice continuous differentiability and
$\nabla_V\ell(V)=0$ yield the following expansion, uniformly over
$\|\epsilon\|_F\leq\rho$ as $\rho\to0$:
\begin{equation}
\ell(V+\epsilon)-\ell(V)=\frac{1}{2}\langle\epsilon,H\epsilon\rangle_F+o(\|\epsilon\|_F^2).
\end{equation}
Writing $\epsilon=\rho U$ with $\|U\|_F\leq1$ gives:
\begin{equation}
\Delta_\rho(V)=\frac{\rho^2}{2}\max_{\|U\|_F\leq1}\langle U,HU\rangle_F+o(\rho^2).
\end{equation}
If $\lambda_{\max}(H)>0$, the maximum is attained by a unit eigenvector associated with $\lambda_{\max}(H)$. If $\lambda_{\max}(H)\leq0$, it is attained at $U=0$. Therefore:
\begin{equation}
\max_{\|U\|_F\leq1}
\langle U,HU\rangle_F=[\lambda_{\max}(H)]_+.
\end{equation}
This proves Eq.~\ref{eq:app_sap_second_order}.
\hfill$\square$

This proposition characterizes the ideal worst-case objective at stationarity, whereas Proposition~3 characterizes the implemented gradient-guided perturbation when $g\neq0$.

\paragraph{Remark.}
Proposition~4 characterizes the ideal worst-case inner problem. At an exact stationary point, the first-order perturbation
$\epsilon^\star=\rho g/\|g\|_F$ is undefined and does not recover the leading Hessian eigenvector. The proposition therefore explains the local geometry of the robust objective rather than claiming that the implemented first-order perturbation solves the stationary second-order problem.
\begin{algorithm}[t]
\caption{TASCO with Random Perturbation}
\label{alg:tasco_random}
\begin{algorithmic}[1]
\REQUIRE Frozen LLM $M_\theta$, test set $\mathcal{D}_{\mathrm{test}}$,
warm-start prefix $V_0$, optimizer $\mathcal{O}$, number of epochs
$E$, perturbation scale $\sigma$, number of perturbations
$K\geq2$, and stability weight $\lambda_{\mathrm{rand}}$
\ENSURE Optimized prefix $V$
\STATE Initialize $V\gets V_0$
\FOR{$e=1,\ldots,E$}
    \STATE Randomly shuffle $\mathcal{D}_{\mathrm{test}}$ and partition it into mini-batches
    \FOR{each mini-batch $\mathcal{B}$}
        \STATE Generate trajectories $\mathbf{Y}_{\mathcal{B}}$ under $V$
        \STATE $\mathcal{L}_{\mathrm{conf}}\gets
        \mathcal{L}_{\mathcal{B}}
        (V;\mathbf{Y}_{\mathcal{B}})$
        \FOR{$k=1,\ldots,K$}
            \STATE Sample
            $\epsilon^{(k)}\sim\mathcal{N}(0,\sigma^2I)$
            and set $V^{(k)}\gets V+\epsilon^{(k)}$
            \STATE Greedily generate trajectories
            $\mathbf{Y}_{\mathcal{B}}^{(k)}$ under $V^{(k)}$
            \STATE Compute
            $\{c_X^{(k)}:X\in\mathcal{B}\}$ using
            Eq.~\ref{eq:trajectory_confidence}
        \ENDFOR
        \STATE Compute $\mathcal{L}_{\mathrm{rand}}$ from
        $\{c_X^{(k)}\}$ using Eq.~\ref{eq:random_sharpness}
        \STATE $g\gets\nabla_V
        \left(
        \mathcal{L}_{\mathrm{conf}}
        +\lambda_{\mathrm{rand}}\mathcal{L}_{\mathrm{rand}}
        \right)$
        \STATE $V\gets\mathcal{O}(V,g)$
    \ENDFOR
\ENDFOR
\STATE \textbf{return} $V$
\end{algorithmic}
\end{algorithm}

\subsection{Relationship Between the Two Local Objectives}

The two analyses reveal distinct views of local geometry. To compare them, suppose temporarily that both perturbation strategies are applied to the same smooth scalar function $h(V)$ and that decoding remains fixed. Away from stationary points, the random-perturbation variance satisfies:
\begin{equation}
\operatorname{Var}_{\epsilon}[h(V+\epsilon)]
=\sigma^2\|\nabla_Vh(V)\|_F^2+\mathcal{O}(\sigma^4).
\end{equation}
By contrast, the worst-case local increase satisfies:
\begin{equation}
\max_{\|\epsilon\|_F\leq\rho}
\left[h(V+\epsilon)-h(V)\right]=\rho\|\nabla_Vh(V)\|_F+\mathcal{O}(\rho^2).
\end{equation}
Thus, the random objective aggregates sensitivity over isotropic directions, whereas the worst-case objective selects the most adverse direction.

At a stationary point with Hessian $H_h$, their leading curvature terms depend on $\|H_h\|_F^2$ and
$[\lambda_{\max}(H_h)]_+$, respectively. Their relationship is bounded by:
\begin{equation}
[\lambda_{\max}(H_h)]_+\leq\|H_h\|_2\leq
\|H_h\|_F.\label{eq:app_curvature_comparison}
\end{equation}
This distinguishes the largest positive curvature from aggregate curvature across directions. In TASCO, Random Perturbation operates on re-decoded trajectory-level log-likelihood, whereas SAP operates on the fixed-trajectory entropy loss. Therefore, these relations clarify their different geometric interpretations but do not establish numerical equivalence or an ordering between the objectives.

% Mean followed by a smaller gray standard deviation.
\newcommand{\meanstd}[2]{%
#1\,{\color{gray}\scriptsize($\pm$#2)}%
}

\begin{table*}[t]
\centering
\setlength{\tabcolsep}{7pt}
\renewcommand{\arraystretch}{1.02}

\begin{NiceTabular}{lcccccc}
\toprule
\textbf{Method}
& \textbf{MATH500}
& \textbf{AMC23}
& \textbf{Minerva}
& \textbf{AIME24}
& \textbf{GPQA}
& \textbf{Avg.} \\
\midrule

\multicolumn{7}{c}{\textbf{Qwen2.5-Math-1.5B}} \\
\midrule
Zero-Shot CoT
& \meanstd{40.2}{1.8}
& \meanstd{34.1}{2.4}
& \meanstd{9.4}{1.4}
& \meanstd{6.2}{3.5}
& \meanstd{19.4}{2.1}
& \meanstd{21.9}{0.5} \\

LTPO
& \meanstd{57.6}{2.1}
& \meanstd{47.2}{2.6}
& \meanstd{12.4}{1.5}
& \meanstd{8.8}{3.1}
& \meanstd{21.5}{1.6}
& \meanstd{28.6}{2.2} \\

TTSV
& \meanstd{69.2}{1.6}
& \meanstd{45.4}{2.0}
& \meanstd{19.2}{1.4}
& \meanstd{6.2}{2.6}
& \meanstd{24.2}{0.6}
& \meanstd{32.9}{1.4} \\

\rowcolor[gray]{0.94}
\textbf{TASCO-RP}
& \meanstd{70.8}{0.4}
& \meanstd{49.9}{0.9}
& \meanstd{22.1}{1.3}
& \meanstd{9.9}{1.0}
& \meanstd{26.7}{1.1}
& \meanstd{35.9}{0.5} \\

\rowcolor[gray]{0.94}
\textbf{TASCO-SAP}
& \meanstd{71.6}{0.9}
& \meanstd{49.4}{1.4}
& \meanstd{20.1}{2.2}
& \meanstd{11.8}{1.6}
& \meanstd{26.2}{0.6}
& \meanstd{36.4}{0.7} \\

\midrule
\multicolumn{7}{c}{\textbf{Qwen2.5-Math-7B}} \\
\midrule
Zero-Shot CoT
& \meanstd{51.8}{1.6}
& \meanstd{42.9}{2.0}
& \meanstd{14.1}{2.9}
& \meanstd{13.3}{2.3}
& \meanstd{30.7}{1.3}
& \meanstd{30.8}{1.8} \\

LTPO
& \meanstd{65.5}{2.2}
& \meanstd{42.8}{3.1}
& \meanstd{18.2}{2.6}
& \meanstd{17.5}{2.8}
& \meanstd{33.6}{2.5}
& \meanstd{35.7}{2.8} \\

TTSV
& \meanstd{70.0}{2.2}
& \meanstd{57.8}{3.7}
& \meanstd{32.8}{2.5}
& \meanstd{17.0}{3.4}
& \meanstd{36.0}{2.5}
& \meanstd{42.7}{3.2} \\

\rowcolor[gray]{0.94}
\textbf{TASCO-RP}
& \meanstd{74.5}{1.2}
& \meanstd{66.7}{2.0}
& \meanstd{38.1}{2.1}
& \meanstd{20.4}{2.4}
& \meanstd{37.7}{2.6}
& \meanstd{47.2}{2.1} \\

\rowcolor[gray]{0.94}
\textbf{TASCO-SAP}
& \meanstd{72.6}{0.8}
& \meanstd{65.2}{2.0}
& \meanstd{39.0}{2.4}
& \meanstd{23.4}{2.4}
& \meanstd{38.5}{2.8}
& \meanstd{47.8}{1.8} \\

\midrule
\multicolumn{7}{c}{\textbf{LLaMA3.1-8B-Instruct}} \\
\midrule
Zero-Shot CoT
& \meanstd{47.0}{0.8}
& \meanstd{25.1}{0.9}
& \meanstd{21.7}{0.7}
& \meanstd{7.9}{1.2}
& \meanstd{28.7}{0.7}
& \meanstd{26.1}{0.8} \\

LTPO
& \meanstd{48.9}{0.8}
& \meanstd{26.5}{1.4}
& \meanstd{20.8}{2.8}
& \meanstd{10.8}{1.4}
& \meanstd{30.1}{0.8}
& \meanstd{27.5}{0.9} \\

TTSV
& \meanstd{47.1}{0.7}
& \meanstd{22.5}{2.8}
& \meanstd{19.6}{2.7}
& \meanstd{5.2}{3.1}
& \meanstd{27.2}{1.7}
& \meanstd{25.0}{1.6} \\

\rowcolor[gray]{0.94}
\textbf{TASCO-RP}
& \meanstd{47.2}{1.2}
& \meanstd{28.6}{0.8}
& \meanstd{25.3}{1.5}
& \meanstd{10.9}{1.2}
& \meanstd{32.7}{0.7}
& \meanstd{29.3}{1.5} \\

\rowcolor[gray]{0.94}
\textbf{TASCO-SAP}
& \meanstd{49.7}{0.8}
& \meanstd{26.4}{2.0}
& \meanstd{26.0}{0.7}
& \meanstd{12.2}{1.2}
& \meanstd{32.2}{0.8}
& \meanstd{29.2}{0.8} \\

\bottomrule
\end{NiceTabular}

\caption{
Accuracy (\%) across three independent random seeds. Each entry reports
the mean accuracy followed by the standard deviation in gray. The average
is computed across the five benchmarks for each run and then aggregated
across seeds.
}
\label{tab:multi_seed_results}
\end{table*}

\paragraph{Summary.}
Together, these analyses show how the two TASCO objectives control complementary forms of local confidence sensitivity. Random Perturbation regularizes distributional variation across perturbed behaviors, whereas SAP controls worst-case sensitivity along an adverse local direction. These results establish local confidence robustness under prefix perturbations rather than a guarantee of answer correctness. Section B separately examines the empirical relationship between local stability and answer accuracy.

\begin{table*}[t]
\centering
\makebox[\textwidth][c]{%
\begin{tabular}{lcccccccccccc}
\toprule
\multirow{2}{*}{\textbf{Method}}
& \multicolumn{2}{c}{\textbf{AIME24}}
& \multicolumn{2}{c}{\textbf{AIME25}}
& \multicolumn{2}{c}{\textbf{MATH500}}
& \multicolumn{2}{c}{\textbf{AMC23}}
& \multicolumn{2}{c}{\textbf{Minerva}}
& \multicolumn{2}{c}{\textbf{Average}}
\\

\cmidrule(lr){2-3}
\cmidrule(lr){4-5}
\cmidrule(lr){6-7}
\cmidrule(lr){8-9}
\cmidrule(lr){10-11}
\cmidrule(lr){12-13}

& \textbf{Acc.} & \textbf{Tok.}
& \textbf{Acc.} & \textbf{Tok.}
& \textbf{Acc.} & \textbf{Tok.}
& \textbf{Acc.} & \textbf{Tok.}
& \textbf{Acc.} & \textbf{Tok.}
& \textbf{Acc.} & \textbf{Tok.}
\\
\midrule

\multicolumn{13}{c}{
\textbf{DeepSeek-R1-Distill-Qwen-1.5B}
}
\\
\midrule

Base
& 22.8 & 7580
& 22.5 & 7674
& 78.6 & 3700
& 59.4 & 6242
& 27.6 & 5057
& 42.2
& 6051
\\

s1
& 25.8 & 7778
& \underline{25.4} & 7982
& 80.2 & 4623
& 61.9 & 6428
& 29.4 & 6127
& 44.5 $\uparrow 2.4$
& 6588 $\uparrow 8.9\%$
\\

CoD
& 26.7 & 7196
& 20.8 & 7552
& 79.4 & \textbf{3267}
& \underline{66.2} & 5293
& 32.0 & \textbf{4041}
& \underline{45.0} $\uparrow 2.8$
& 5470 $\downarrow 9.6\%$
\\

$\alpha1$
& \textbf{30.8} & \underline{6841}
& 12.9 & \textbf{6529}
& \underline{81.0} & \underline{3541}
& 63.1 & \underline{4310}
& \underline{33.1} & 4589
& 44.2 $\uparrow 2.0$
& \textbf{5162} $\downarrow 14.7\%$
\\

\rowcolor[gray]{0.94}
\textbf{TASCO-SAP}
& \underline{30.4} & \textbf{6751}
& \textbf{29.2} & \underline{6838}
& \textbf{81.4} & 3616
& \textbf{69.4} & \textbf{4282}
& \textbf{34.6} & \underline{4507}
& \textbf{49.0} $\uparrow 6.8$
& \underline{5199} $\downarrow 14.1\%$
\\

\midrule

\multicolumn{13}{c}{
\textbf{DeepSeek-R1-Distill-Qwen-7B}
}
\\
\midrule

Base
& 44.2 & 6899
& 25.0 & 7099
& 89.4 & 3399
& 79.4 & 4672
& \underline{42.6} & 4416
& 56.1
& 5297
\\

s1
& 37.5 & 7368
& 24.2 & 7041
& 89.0 & 4072
& 78.1 & 5271
& 40.4 & 4986
& 53.8 $\downarrow 2.3$
& 5748 $\uparrow 8.5\%$
\\

CoD
& \underline{45.0} & \underline{6301}
& \underline{33.3} & \textbf{6582}
& 88.0 & \textbf{2071}
& 85.6 & \textbf{3645}
& 40.1 & \textbf{2440}
& \underline{58.4} $\uparrow 2.3$
& \textbf{4208} $\downarrow 20.6\%$
\\

$\alpha1$
& 36.7 & 6831
& 32.5 & \underline{6638}
& \underline{89.8} & 3848
& \textbf{90.6} & 4529
& \underline{42.6} & 4223
& \underline{58.4} $\uparrow 2.3$
& 5214 $\downarrow 1.6\%$
\\

\rowcolor[gray]{0.94}
\textbf{TASCO-SAP}
& \textbf{49.2} & \textbf{6188}
& \textbf{34.2} & 6652
& \textbf{90.4} & \underline{3041}
& \underline{88.1} & \underline{4287}
& \textbf{43.4} & \underline{3988}
& \textbf{61.1} $\uparrow 4.9$
& \underline{4831} $\downarrow 8.8\%$
\\

\bottomrule
\end{tabular}%
}
\caption{
Accuracy (\%) and average generation length across five reasoning benchmarks.
Accuracy is reported to one decimal place, and token counts are rounded to the
nearest integer. Average results are computed across the five datasets. Arrows
denote absolute accuracy changes and relative token changes compared with Base.
Best results are in \textbf{bold}, and second-best results are
\underline{underlined}.
}
\label{tab:reasoning_accuracy_efficiency}
\end{table*}

\section{D. Complete Optimization Algorithms}
\label{app:algorithms}

This section provides the complete optimization procedures for the two TASCO variants. Both algorithms optimize a task-level prefix $V$ over the unlabeled test set while keeping the LLM parameters $\theta$ frozen. They differ only in how local confidence stability is estimated and incorporated into prefix optimization.

\paragraph{Random Perturbation.}
Algorithm~\ref{alg:tasco_random} presents the Random Perturbation variant. At each update, it samples $K$ Gaussian perturbations around the current prefix and generates a trajectory under each perturbed prefix. The resulting trajectory-level confidence variance is averaged across the mini-batch and combined with the confidence loss to update $V$.

\paragraph{Sharpness-Aware Perturbation.}
Algorithm~\ref{alg:tasco_sap} presents the Sharpness-Aware Perturbation variant. It first generates one trajectory per input under the current prefix and holds these trajectories fixed throughout the update. A gradient-guided perturbation is then constructed from the confidence loss, after which the prefix is updated using the loss evaluated at the perturbed prefix. The perturbation is treated as constant during the second backward pass.

\paragraph{Computational Cost.}
Random Perturbation captures richer distributional information but requires $K$ independently decoded perturbed rollouts per input, introducing additional adaptation cost. Sharpness-Aware Perturbation avoids these rollouts through a single gradient-guided perturbation on the fixed trajectory, substantially reducing autoregressive decoding cost relative to Random Perturbation. The reported token-efficiency results characterize generation length after adaptation. Moreover, the learned prefix transfers to unseen datasets without re-optimization, avoiding additional adaptation cost in cross-distribution settings.

\begin{algorithm}[t]
\caption{TASCO with Sharpness-Aware Perturbation}
\label{alg:tasco_sap}
\begin{algorithmic}[1]
\REQUIRE Frozen LLM $M_\theta$, test set $\mathcal{D}_{\mathrm{test}}$,
warm-start prefix $V_0$, optimizer $\mathcal{O}$, number of epochs
$E$, and perturbation radius $\rho$
\ENSURE Optimized prefix $V$
\STATE Initialize $V\gets V_0$
\FOR{$e=1,\ldots,E$}
    \STATE Randomly shuffle $\mathcal{D}_{\mathrm{test}}$ and partition it into mini-batches
    \FOR{each mini-batch $\mathcal{B}$}
        \STATE Generate trajectories $\mathbf{Y}_{\mathcal{B}}$ under $V$ and hold them fixed
        \STATE $g\gets\nabla_V
        \mathcal{L}_{\mathcal{B}}
        (V;\mathbf{Y}_{\mathcal{B}})$
        \IF{$\|g\|_F>0$}
            \STATE $\epsilon^\star\gets
            \rho g/\|g\|_F$
        \ELSE
            \STATE $\epsilon^\star\gets0$
        \ENDIF
        \STATE $g_{\mathrm{SAP}}\gets
        \nabla_V
        \mathcal{L}_{\mathcal{B}}
        \left(
        V+\operatorname{StopGrad}(\epsilon^\star);
        \mathbf{Y}_{\mathcal{B}}
        \right)$
        \STATE $V\gets\mathcal{O}(V,g_{\mathrm{SAP}})$
    \ENDFOR
\ENDFOR
\STATE \textbf{return} $V$
\end{algorithmic}
\end{algorithm}

\section{E. Further Analysis of Reasoning Effectiveness}
\label{app:further_reasoning_analysis}

This section examines TASCO beyond final-answer accuracy, focusing on whether its gains remain consistent across models and runs and are reflected in the reasoning process itself. We also assess its sensitivity to the perturbation scale.

\subsection{Evaluation on Reasoning-Enhanced Models.}
We first examine whether TASCO also benefits reasoning-enhanced LLMs, whose pretrained reasoning behavior differs substantially from that of general instruction-tuned models. We evaluate DeepSeek-R1-Distill-Qwen-1.5B and DeepSeek-R1-Distill-Qwen-7B on five mathematical reasoning benchmarks. In addition to the original model, we compare TASCO with test-time reasoning-control methods, including s1, CoD, and $\alpha1$. We report answer accuracy and average generation length to jointly assess reasoning effectiveness and token efficiency.

As shown in Table~\ref{tab:reasoning_accuracy_efficiency}, TASCO achieves the highest average accuracy at both model scales and consistently outperforms the base model across all five benchmarks. It improves average accuracy from 42.2\% to 49.0\% on DeepSeek-R1-Distill-Qwen-1.5B while reducing generation length by 14.1\%, and from 56.1\% to 61.1\% on the 7B model with an 8.8\% reduction. Although some reasoning-control baselines produce shorter outputs, their accuracy gains are smaller or inconsistent. TASCO thus offers a stronger balance between accuracy and generation efficiency, demonstrating that stability-aware confidence optimization also benefits models already trained for deliberative reasoning.

\subsection{Evaluation Across Random Seeds}
\label{app:multi_seed_evaluation}

We additionally evaluate TASCO and selected baselines using three independent random seeds. To limit the cost of repeated test-time optimization, we include representative methods covering no adaptation, latent-variable optimization, and confidence-only prefix optimization: Zero-Shot CoT, LTPO, and TTSV. We evaluate both TASCO variants under the same protocol and report mean accuracy with standard deviation in Table~\ref{tab:multi_seed_results}. This evaluation provides more reliable performance estimates and quantifies variation across independent runs.

\begin{table*}[t]
\centering
\small
\setlength{\tabcolsep}{5pt}
\begin{tabular}{cccccccccc}
\toprule
\multirow{2}{*}{Model}
& \multirow{2}{*}{Method}
& \multicolumn{4}{c}{MATH500}
& \multicolumn{4}{c}{Minerva Math} \\
\cmidrule(lr){3-6}\cmidrule(lr){7-10}
& &
SC \(\uparrow\)
& SI \(\uparrow\)
& PC \(\downarrow\)
& RR \(\downarrow\)
& SC \(\uparrow\)
& SI \(\uparrow\)
& PC \(\downarrow\)
& RR \(\downarrow\) \\
\midrule

\multirow{2}{*}{Qwen2.5-Math-7B}
& Confidence Only
& 82.7 & 77.6 & 18.8 & 15.1
& 75.4 & 75.7 & 22.8 & 11.2 \\
& TASCO-SAP
& 85.2 & 80.4 & 17.2 & 11.6
& 77.5 & 80.5 & 19.5 & 10.1 \\

\midrule

\multirow{2}{*}{LLaMA3.1-8B-Instruct}
& Confidence Only
& 65.6 & 66.2 & 43.0 & 10.2
& 51.8 & 58.8 & 51.8 & 12.8 \\
& TASCO-SAP
& 67.6 & 67.6 & 41.4 & 8.8
& 63.7 & 69.6 & 40.4 & 10.2 \\

\midrule

\multirow{2}{*}{DS-Distill-Qwen-7B}
& Confidence Only
& 91.2 & 83.6 & 6.8 & 11.6
& 69.5 & 69.7 & 31.3 & 11.9 \\
& TASCO-SAP
& 94.2 & 85.1 & 4.8 & 11.9
& 72.1 & 72.6 & 29.3 & 10.6 \\

\bottomrule
\end{tabular}
\caption{Reasoning-process evaluation using DeepSeek-V4-Flash as the external judge. All values are percentages.}
\label{tab:reasoning_process_evaluation}
\end{table*}

\subsection{Semantic Evaluation of Reasoning Processes}
\label{app:reasoning_process_evaluation}

\paragraph{Experimental Setup.}
We use DeepSeek-V4-Flash as an external judge to examine how TASCO affects intermediate reasoning. We compare paired trajectories generated by Confidence Only and TASCO-SAP using Qwen2.5-Math-7B, LLaMA3.1-8B-Instruct, and DeepSeek-R1-Distill-Qwen-7B on the full MATH500 and Minerva Math test sets, containing 500 and 272 examples, respectively. For each trajectory, the judge receives only the original problem and the generated reasoning process; no ground-truth answer is provided. Common explicit final-answer spans are masked to prevent the judge from inferring reasoning quality from answer correctness. We enable thinking mode with high reasoning effort and request a structured JSON annotation.

The judge first divides each trajectory into semantically meaningful steps. Each step receives a correctness label from \{\texttt{correct}, \texttt{incorrect}, \texttt{uncertain}\}, an informativeness label from \{\texttt{informative}, \texttt{redundant}, \texttt{off\_track}\}, and a functional role. The judge also determines whether the trajectory commits prematurely to an answer or reasoning direction before sufficient support has been established. All reported metrics are computed from these annotations rather than directly assigned by the judge.

\paragraph{External-Judge Prompt.}
The prompt template used for reasoning-process annotation is shown below. The placeholders are replaced with the original problem and the corresponding trajectory after final-answer masking.

\paragraph{External-Judge Prompt.}
We use the following condensed prompt to summarize the instructions given to the external judge. The complete prompt and JSON schema are provided in the released implementation.

\begin{center}
\begin{tcolorbox}[
  enhanced,
  breakable,
  colback=gray!6,
  colframe=gray!65,
  width=0.98\linewidth,
  arc=1mm,
  boxsep=1mm,
  top=1.5mm,
  bottom=1.5mm,
  left=2mm,
  right=2mm,
  title={Reasoning-Process Evaluation Prompt},
  fonttitle=\bfseries,
  fontupper=\small
]

\textbf{System:}
You are an expert evaluator of mathematical reasoning. Evaluate the reasoning process independently of its final answer, writing style, verbosity, or expressed confidence.

Divide the trajectory into semantically meaningful steps. Label each step by:

\begin{itemize}
  \setlength{\itemsep}{0pt}
  \setlength{\parskip}{0pt}
  \item correctness: \texttt{correct}, \texttt{incorrect}, or \texttt{uncertain};
  \item informativeness: \texttt{informative}, \texttt{redundant}, or \texttt{off\_track}.
\end{itemize}

Determine whether the trajectory commits prematurely to an answer or reasoning direction before sufficient support is established. Return all annotations as a valid JSON object.

\smallskip
\textbf{User:}
Evaluate the following reasoning trajectory.

\smallskip
\textbf{Problem:}
\textcolor{blue!70!black}{\texttt{<Problem>}}

\smallskip
\textbf{Reasoning trajectory:}
\textcolor{blue!70!black}{\texttt{<Reasoning Trajectory>}}

\end{tcolorbox}
\end{center}

\begin{figure*}[t]
    \centering
    \includegraphics[width=\textwidth]{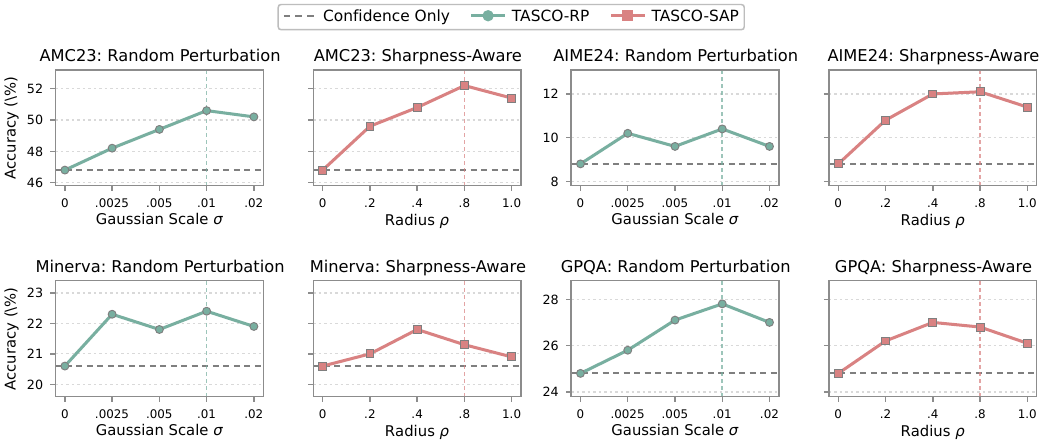}
    \caption{Sensitivity of TASCO to perturbation scales on four additional benchmarks using Qwen2.5-Math-1.5B. Each benchmark contains separate panels for Random Perturbation and Sharpness-Aware Perturbation.}
    \label{fig:additional_scale_sensitivity}
\end{figure*}

\paragraph{Evaluation Metrics.}
For trajectory $i$ with $J_i$ semantic steps, let $c_{ij}$ and $u_{ij}$ denote the correctness and informativeness labels of step $j$, respectively. Let $E_i$ denote the subset of steps labeled either correct or incorrect. Step Correctness is defined as:
\begin{equation}
\mathrm{SC}_i
=
\frac{
\sum_{j\in E_i}
\mathbf{1}\!\left\{c_{ij}=\mathrm{correct}\right\}
}{
|E_i|
},
\label{eq:step_correctness}
\end{equation}
where $\mathbf{1}\{\cdot\}$ is the indicator function and uncertain steps are excluded. Higher SC indicates that a larger proportion of assessable steps are mathematically or logically valid.

Step Informativeness measures the proportion of steps that introduce useful information and advance the solution:
\begin{equation}
\mathrm{SI}_i
=
\frac{1}{J_i}
\sum_{j=1}^{J_i}
\mathbf{1}\!\left\{u_{ij}=\mathrm{informative}\right\}.
\label{eq:step_informativeness}
\end{equation}
Higher SI indicates more productive reasoning.

Redundancy Rate measures the proportion of steps that mainly repeat or restate previously established information:
\begin{equation}
\mathrm{RR}_i
=
\frac{1}{J_i}
\sum_{j=1}^{J_i}
\mathbf{1}\!\left\{u_{ij}=\mathrm{redundant}\right\}.
\label{eq:redundancy_rate}
\end{equation}
Lower RR indicates less repetitive reasoning.

Let $p_i$ denote the trajectory-level premature-commitment judgment. We define the corresponding indicator and its aggregate rate over $N$ trajectories as:
\begin{equation}
\mathrm{PC}_i
=
\mathbf{1}\!\left\{p_i=\mathrm{true}\right\},
\qquad
\mathrm{PC}
=
\frac{1}{N}\sum_{i=1}^{N}\mathrm{PC}_i.
\label{eq:premature_commitment}
\end{equation}
Lower PC indicates that the model is less likely to commit prematurely to an insufficiently supported reasoning direction. We average SC, SI, and RR across trajectories and report all four metrics as percentages.

\paragraph{Sensitivity to Perturbation Scale.}
We further examine the sensitivity of TASCO to the Gaussian scale $\sigma$ and perturbation radius $\rho$ on four additional benchmarks. As shown in Fig.~\ref{fig:additional_scale_sensitivity}, both variants outperform confidence-only optimization across a broad range of perturbation scales. The default settings, $\sigma=0.01$ and $\rho=0.8$, achieve the best or near-best accuracy across datasets, while moderate changes generally produce comparable results. These findings indicate that TASCO is not sensitive to a narrowly tuned perturbation scale and that the selected settings generalize well across benchmarks.

\paragraph{Results.}
As shown in Table~\ref{tab:reasoning_process_evaluation}, TASCO consistently improves all four reasoning-process metrics across both datasets. It produces a larger proportion of correct and informative steps while reducing redundant reasoning and premature commitment. These results suggest that stability-aware optimization affects not only final-answer accuracy but also how the model develops and maintains its reasoning process.

\section{F. Case Studies}
\label{app:case_studies}

This section presents case studies that illustrate the experimental prompting protocol and compare the reasoning behaviors induced by Confidence Only and TASCO.

\paragraph{Prompt and Example.}
We first provide an AIME24 example illustrating the complete prompt and response format. As shown in Table~\ref{tab:case_aime24}, the system message instructs the model to reason step by step and place its final answer within \texttt{\textbackslash boxed\{\}}, while the original problem is provided as the user message. TASCO adds no textual instructions; its learned continuous prefix is prepended only at the embedding level. The table reports the complete response and the final answer extracted by the benchmark parser. Subsequent case studies use the same prompt format and omit these repeated instructions.

\paragraph{Reasoning Patterns.}
Table~\ref{tab:three_case_comparison} reveals a recurring contrast between the two methods. Confidence Only often commits early to a locally plausible solution and continues confidently after discarding a necessary condition. Across the three cases, it cancels a potentially zero factor, confuses simultaneous constraints with their union, or violates a coprimality requirement. TASCO instead preserves these constraints throughout the derivation and retains valid reasoning branches until they are properly resolved. These examples suggest that stability-aware optimization mitigates premature commitment to fragile reasoning paths and promotes more consistent and reliable multi-step reasoning.
\begin{table*}[p]
\centering
\begin{minipage}{0.94\textwidth}
\raggedright
\hrule

\vspace{6pt}

\textcolor{teal}{\textbf{SYSTEM:}}
Please reason step by step, and put your final answer within
\texttt{\textbackslash boxed\{\}}.

\vspace{6pt}

\textcolor{teal}{\textbf{USER:}}
\textcolor{teal!75!black}{
\([\mathbf{v}_1][\mathbf{v}_2][\mathbf{v}_3]\cdots[\mathbf{v}_{20}]\)
}
Every morning Aya goes for a \(9\)-kilometer-long walk and stops
at a coffee shop afterwards. When she walks at a constant speed
of \(s\) kilometers per hour, the walk takes her \(4\) hours,
including \(t\) minutes spent in the coffee shop. When she walks
at \(s+2\) kilometers per hour, the walk takes her \(2\) hours and
\(24\) minutes, including \(t\) minutes spent in the coffee shop.
Suppose Aya walks at \(s+\frac{1}{2}\) kilometers per hour. Find
the number of minutes the walk takes her, including the \(t\)
minutes spent in the coffee shop.

\vspace{6pt}

\textcolor{teal}{\textbf{GROUND TRUTH:}}
204

\vspace{6pt}

\textcolor{green!50!black}{\textbf{TASCO REASONING:}}

\vspace{3pt}

To solve this problem, we first determine the values of \(s\) and
\(t\) using the two given walking times.

\vspace{4pt}

\textbf{\#\# Step 1: Formulate the first time constraint.}
When Aya walks at \(s\) kilometers per hour, the walking time is
\(\frac{9}{s}\) hours, while the time spent in the coffee shop is
\(\frac{t}{60}\) hours. Since the total time is \(4\) hours,
\[
\frac{9}{s}+\frac{t}{60}=4.
\]

\textbf{\#\# Step 2: Formulate the second time constraint.}
When Aya walks at \(s+2\) kilometers per hour, the total duration
is \(2\) hours and \(24\) minutes, or \(2.4\) hours. Therefore,
\[
\frac{9}{s+2}+\frac{t}{60}=2.4.
\]

\textbf{\#\# Step 3: Solve for \(s\) and \(t\).}
Solving the two equations gives
\[
(s,t)=(-3,240)
\qquad\text{or}\qquad
(s,t)=(2.5,24).
\]
Because a speed of \(-3\) kilometers per hour is not physically
meaningful, we discard the first solution. Hence,
\[
s=2.5,\qquad t=24.
\]

\textbf{\#\# Step 4: Compute the requested duration.}
At the new speed,
\[
s+\frac{1}{2}=2.5+0.5=3
\]
kilometers per hour. Walking \(9\) kilometers therefore takes
\[
\frac{9}{3}=3
\]
hours. Including the \(24\) minutes spent in the coffee shop, the
total duration is
\[
3\times60+24=204
\]
minutes.

Therefore, the walk takes \(\boxed{204}\) minutes.

\vspace{6pt}
\hrule
\end{minipage}

\caption{Prompt and TASCO reasoning example on AIME24. The vectors shown before the user question schematically represent the learned continuous prefix prepended at the embedding level rather than textual tokens.}
\label{tab:case_aime24}
\end{table*}

\clearpage
\onecolumn

\setlength{\LTleft}{0pt}
\setlength{\LTright}{0pt}
\setlength{\LTpre}{6pt}
\setlength{\LTpost}{10pt}
\renewcommand{\arraystretch}{1.08}

\begin{longtable}{@{}p{\textwidth}@{}}
\toprule

\textcolor{teal}{\textbf{CASE 1: POLYNOMIAL ROOT FINDING}} \\[5pt]

\textcolor{teal}{\textbf{QUESTION:}}
Find the roots of
\[
(x-3)^3+(x-7)^3=(2x-10)^3.
\]
\\[-2pt]

\textcolor{teal}{\textbf{GROUND TRUTH:}}
\(3,5,7\) \\[4pt]

\textcolor{red}{\textbf{CONFIDENCE-ONLY REASONING (INCORRECT):}}
\\[2pt]

\textbf{\#\# Step 1:}
To solve the equation, we use the identity
\[
a^3+b^3=(a+b)(a^2-ab+b^2).
\]

\textbf{\#\# Step 2:}
Let \(a=x-3\) and \(b=x-7\). Applying the identity gives
\[
(2x-10)
\bigl((x-3)^2-(x-3)(x-7)+(x-7)^2\bigr)
=(2x-10)^3.
\]

\textbf{\#\# Step 3:}
\textcolor{red}{Dividing both sides by \(2x-10\)}, we obtain
\[
(x-3)^2-(x-3)(x-7)+(x-7)^2=(2x-10)^2.
\]

\textbf{\#\# Step 4:}
Expanding both sides gives
\[
x^2-6x+9-(x^2-10x+21)+x^2-14x+49
=4x^2-40x+100.
\]
Simplifying yields
\[
x^2-10x+37=4x^2-40x+100.
\]

\textbf{\#\# Step 5:}
Rearranging and factoring gives
\[
3x^2-30x+63=0,
\qquad
x^2-10x+21=(x-3)(x-7)=0.
\]

\textbf{\#\# Step 6:}
Therefore, the roots are \(x=3\) and \(x=7\), and the final answer is \{3,7\}.

\textcolor{green!50!black}{\textbf{TASCO REASONING (CORRECT):}}
\\[3pt]

\textbf{\#\# Step 1:}
Let \(y=x-5\). Then
\[
x-3=y+2,\qquad x-7=y-2,\qquad 2x-10=2y.
\]

\textbf{\#\# Step 2:}
The equation becomes
\[
(y+2)^3+(y-2)^3=(2y)^3.
\]
Expanding both sides gives
\[
(y^3+6y^2+12y+8)
+(y^3-6y^2+12y-8)=8y^3.
\]

\textbf{\#\# Step 3:}
Combining like terms, rearranging, and factoring gives
\[
2y^3+24y=8y^3,
\qquad
6y(y^2-4)=0.
\]

\textbf{\#\# Step 4:}
Thus,
\[
y=0,\qquad y=2,\qquad y=-2.
\]
Substituting \(x=y+5\) gives
\[
x=5,\qquad x=7,\qquad x=3.
\]

\textbf{\#\# Step 5:}
Therefore, the complete set of roots is \{3,5,7\}.

\\[4pt]

\midrule

\textcolor{teal}{\textbf{CASE 2: DOMAIN CONSTRAINTS}} \\[5pt]

\textcolor{teal}{\textbf{QUESTION:}}
For which real values of \(x\) is the following function undefined?
\[
f(x)=\frac{1}{|x^2+3x-4|+|x^2+9x+20|}.
\]
\\[-2pt]

\textcolor{teal}{\textbf{GROUND TRUTH:}}
\(-4\) \\[4pt]

\textcolor{red}{\textbf{CONFIDENCE-ONLY REASONING (INCORRECT):}}
\\[3pt]

\textbf{\#\# Step 1:}
To determine the values outside the domain, we find where the
denominator is zero. Since both terms are absolute values,
\[
|x^2+3x-4|+|x^2+9x+20|=0
\]
if and only if both quadratic expressions are zero.

\textbf{\#\# Step 2:}
The first quadratic factors as
\[
x^2+3x-4=(x+4)(x-1)=0,
\]
giving \(x=-4\) or \(x=1\).

\textbf{\#\# Step 3:}
The second quadratic factors as
\[
x^2+9x+20=(x+4)(x+5)=0,
\]
giving \(x=-4\) or \(x=-5\).

\textbf{\#\# Step 4:}
\textcolor{red}{The roots of the two equations are combined by
union rather than intersection.}
The reasoning therefore concludes that the denominator is zero at
\(x=-5,-4,1\), giving \{-5,-4,1\}.

\textcolor{green!50!black}{\textbf{TASCO REASONING (CORRECT):}}
\\[3pt]

\textbf{\#\# Step 1:}
The denominator is a sum of two nonnegative terms, so it is zero
only when both quadratic expressions are zero simultaneously.

\textbf{\#\# Step 2:}
Solving the first equation gives
\[
x^2+3x-4=(x+4)(x-1)=0,
\]
whose roots are \(x=-4\) and \(x=1\).

\textbf{\#\# Step 3:}
Solving the second equation gives
\[
x^2+9x+20=(x+4)(x+5)=0,
\]
whose roots are \(x=-4\) and \(x=-5\).

\textbf{\#\# Step 4:}
Because the two conditions must hold simultaneously, we take the
intersection:
\[
\{-4,1\}\cap\{-4,-5\}=\{-4\}.
\]
Therefore, the denominator is zero only at \(x=-4\), and the final
answer is \{-4\}.
\\[4pt]

\midrule

\textcolor{teal}{\textbf{CASE 3: COPRIMALITY CONSTRAINTS}} \\[5pt]

\textcolor{teal}{\textbf{QUESTION:}}
If \(a\) and \(b\) are positive integers such that
\(\gcd(a,b)=210\),
\(\operatorname{lcm}(a,b)=210^3\), and \(a<b\), how many possible
values are there for \(a\)?
\\[5pt]

\textcolor{teal}{\textbf{GROUND TRUTH:}}
\(8\) \\[4pt]

\textcolor{red}{\textbf{CONFIDENCE-ONLY REASONING (INCORRECT):}}
\\[3pt]

\textbf{\#\# Step 1:}
Using
\[
\gcd(a,b)\operatorname{lcm}(a,b)=ab,
\]
we obtain
\[
ab=210\cdot210^3=210^4.
\]

\textbf{\#\# Step 2:}
Since \(\gcd(a,b)=210\), write
\[
a=210m,\qquad b=210n,
\]
where \(\gcd(m,n)=1\). Substituting gives
\[
mn=210^2.
\]

\textbf{\#\# Step 3:}
The prime factorization is
\[
210^2=2^2\cdot3^2\cdot5^2\cdot7^2.
\]
Therefore, the total number of divisors of \(210^2\) is
\[
(2+1)^4=81.
\]

\textbf{\#\# Step 4:}
\textcolor{red}{Each divisor is treated as producing a coprime
factor pair, even though many divisor pairs share prime factors.}
After removing symmetric pairs, the number of possible values of
\(a\) is computed as
\[
\frac{81-1}{2}=40.
\]
Therefore, the final answer is \{40\}.

\textcolor{green!50!black}{\textbf{TASCO REASONING (CORRECT):}}
\\[3pt]

\textbf{\#\# Step 1:}
Using the relationship between the greatest common divisor and
least common multiple gives
\[
ab=\gcd(a,b)\operatorname{lcm}(a,b)
=210\cdot210^3=210^4.
\]

\textbf{\#\# Step 2:}
Write
\[
a=210m,\qquad b=210n,
\]
where \(\gcd(m,n)=1\). Then
\[
(210m)(210n)=210^4,
\qquad
mn=210^2.
\]

\textbf{\#\# Step 3:}
Since
\[
210^2=2^2\cdot3^2\cdot5^2\cdot7^2
\]
and \(m\) and \(n\) are coprime, each complete prime power must be
assigned entirely to either \(m\) or \(n\). Splitting a prime
power between the two factors would violate \(\gcd(m,n)=1\).

\textbf{\#\# Step 4:}
There are four distinct prime powers, each of which can be
assigned independently to either \(m\) or \(n\). Hence, there are
\[
2^4=16
\]
ordered assignments. Since \(a<b\), equivalently \(m<n\), each
unordered pair contributes exactly one valid value of \(a\).
Therefore,
\[
\frac{16}{2}=8.
\]
The final answer is \{8\}.
\\[4pt]

\bottomrule
\caption{Three comparative case studies on polynomial roots, domain constraints, and coprime factorization. In each case, Confidence Only drops a condition identified earlier in the derivation, whereas TASCO preserves the condition and obtains the correct answer.}
\label{tab:three_case_comparison}
\\
\end{longtable}

\end{document}